\documentclass[final,5p,times,twocolumn,authoryear]{elsarticle}

\usepackage{amssymb}
\usepackage{amsthm}
\usepackage{amsmath}
\usepackage{enumerate}
\usepackage{booktabs}
\usepackage{multirow}
\usepackage{color}
\newtheorem{theorem}{Theorem}
\newtheorem{corollary}{Corollary}
\newtheorem{lemma}{Lemma}
\newtheorem{definition}{Definition}

\newtheorem{example}{Example}

\newcommand{\R}{\mathbb{R}}
\DeclareMathOperator*{\argmax}{arg\,max}

\newcommand{\norm}[1]{\left\lVert#1\right\rVert}

\journal{ }
\date{}

\begin{document}

\begin{frontmatter}



\title{SympNets: Intrinsic structure-preserving symplectic networks for identifying Hamiltonian systems}


\author[cas,ucas]{Pengzhan Jin\fnref{cof}}
\author[brown]{Zhen Zhang\fnref{cof}}
\author[cas,ucas]{Aiqing Zhu}
\author[cas,ucas]{Yifa Tang\corref{cor}}
\author[brown]{George Em Karniadakis\corref{cor}}
\cortext[cor]{Corresponding authors:\\ \hspace*{3.5ex} tyf@lsec.cc.ac.cn (Yifa Tang);\\ \hspace*{3.5ex} george\_karniadakis@brown.edu (George Em Karniadakis)}
\fntext[cof]{Pengzhan Jin and Zhen Zhang contributed equally to this work.}

\address[cas]{LSEC, ICMSEC, Academy of Mathematics and Systems Science, Chinese Academy of Sciences, Beijing 100190, China}
\address[ucas]{School of Mathematical Sciences, University of Chinese Academy of Sciences, Beijing 100049, China}
\address[brown]{Division of Applied Mathematics, Brown University, Providence, RI 02912, USA}

\begin{abstract}
We propose new symplectic networks (SympNets) for identifying Hamiltonian systems from data based on a composition of linear, activation and gradient modules. In particular, we define two classes of SympNets: the LA-SympNets composed of linear and activation modules, and the G-SympNets composed of gradient modules. Correspondingly, we prove two new universal approximation theorems that demonstrate that SympNets can approximate arbitrary symplectic maps based on appropriate activation functions. We then perform several experiments including the pendulum, double pendulum and three-body problems to investigate the expressivity and the generalization ability of SympNets. The simulation results show that even very small size SympNets can generalize well, and are able to handle both separable and non-separable Hamiltonian systems with data points resulting from short or long time steps. In all the test cases, SympNets outperform the baseline models, and are much faster in training and prediction. We also develop an extended version of SympNets to learn the dynamics from irregularly sampled data. This extended version of SympNets can be thought of as a universal model representing the solution to an arbitrary Hamiltonian system.
\end{abstract}




\end{frontmatter}


\section{Introduction}
\label{intro}

It is well known that neural networks can approximate continuous maps \citep{cybenko1989approximation,hornik1989multilayer}. However, universal approximation theorems only guarantee a small approximation error for a sufficiently large network, but do not consider the optimization and generalization errors. In order to obtain satisfactory results for a given task,  big data is required that we may not be able to afford if this task is regarded as a pure approximation problem \citep{jin2019quantifying}. For this reason, when applying deep learning to  physical systems, the cost of data acquisition is prohibitive, and we are inevitably faced with the challenge of drawing conclusions and making decisions under partial information. Fortunately, for physical systems there exists a vast amount of prior knowledge that is not always utilized in machine learning practices. Encoding such structured information into a learning algorithm results in amplifying substantially the information content of the data that the algorithm sees, enabling it to quickly steer itself towards the right solution and to generalize well even when only a few training examples are available \citep{lagaris1998artificial, raissi2019physics}. There have been many research works focusing on how to employ prior knowledge to construct the targeted machine learning algorithms for specific problems, where the approximated maps usually have special structures or properties, which we naturally expect the trained networks to possess, such as image classification \citep{krizhevsky2012imagenet}, natural language processing \citep{maas2013rectifier}, game playing \citep{silver2016mastering}, as well as the recent work \citep{lu2019deeponet} providing a special network structure based on the universal approximation theorem for approximating nonlinear operators \citep{chen1995universal}. Some works enforce the prior information on network structures by manifold calculus \citep{fiori2008lie, fiori2011numerical, fiori2011theoretical}. Additionally, more contributions specific to solving problems on the manifold of symplectic matrices
were proposed in~\cite{fiori2016riemannian, fiori2017exact, wang2018riemannian}.

In this work, we aim to study how to impose the prior information on the neural networks for identifying Hamiltonian systems. Specifically,
we focus on endowing the neural networks with a symplectic structure.

First, we provide some relevant background material. Denote the $d$-by-$d$ identity matrix by $I_{d}$, and let
\begin{equation*}
J:=\begin{pmatrix} 0 & I_{d} \\ -I_{d} & 0 \end{pmatrix},
\end{equation*}
which is an orthogonal, skew-symmetric real matrix, so that $J^{-1}=J^{T}=-J$.
\begin{definition}
A matrix $H\in \R^{2d\times2d}$ is called symplectic if $H^{T}JH=J$.
\end{definition}
With the concept of symplectic matrix, the definition of symplectic map can be given.
\begin{definition} \label{def:symp_map}
A differentiable map $\Phi:U\rightarrow\R^{2d}$ (where $U\subset\R^{2d}$ is an open set) is called symplectic if the Jacobian matrix $\frac{\partial \Phi}{\partial x}$ is everywhere symplectic, i.e.,
\begin{equation*}
    \left(\frac{\partial \Phi}{\partial x}\right)^T J \left(\frac{\partial \Phi}{\partial x}\right) = J.
\end{equation*}
\end{definition}

We consider the Hamiltonian system
\begin{equation} \label{eq:hm_sys}
\left\{\begin{aligned}
&\dot{y}=J^{-1}\nabla H(y) \\
&y(t_{0})=y_{0}
\end{aligned}\right.,
\end{equation}
where $y(t)\in \R^{2d}$, and $H$ is the Hamiltonian typically representing the energy of the system (\ref{eq:hm_sys}). Let $\phi_{t}(y_{0})$ be the phase flow of system (\ref{eq:hm_sys}). In 1899, Poincare pointed out that the phase flow of a Hamiltonian system is a symplectic map \citep[p. 184, Theorem 2.4]{hairer2006geometric}, i.e.,
\begin{equation} \label{eq:sym_con}
\left(\frac{\partial \phi_t}{\partial y_0}\right)^T J \left(\frac{\partial \phi_t}{\partial y_0}\right) = J.
\end{equation}
The evaluation of the behavior of dynamical systems at long time is a notoriously difficult problem in mathematics, particularly for discrete dynamical systems. One may encounter situations where the dynamics explodes, converges to stationary states or exhibits chaotic behavior. Fortunately, for Hamiltonian systems, these problems can be alleviated by imposing the symplectic structure on the numerical methods due to (\ref{eq:sym_con}). There are some well-developed works on symplectic integration, see for example \citep{feng1984difference,hairer2006geometric,lubich2008quantum}. As the symplectic numerical integrators yield transformative results across diverse applications based on the Hamiltonian systems \citep{omelyan2003symplectic,faou2009computing,zhang2014canonicalization,qin2015canonical}, we aim to consider the construction of networks possessing symplecticity and explore how it impacts the numerical methods for the Hamiltonian systems.

To this end, many neural network-based models have been proposed to identify the Hamiltonian systems from data \citep{tom2019ham,greydanus2019hamiltonian,rezende2019equivariant,sanchez2019hamiltonian,chen2020symplectic, Toth2020Hamiltonian, Zhong2020Symplectic}, with further applications in image prediction \citep{greydanus2019hamiltonian}, generative modeling \citep{Toth2020Hamiltonian} and continuous control \citep{Zhong2020Symplectic}. These learning models are mostly constructed by exploiting the structure of standard numerical time-stepping methods \citep{yannis1998rk, chen2018neural, raissi2018multistep}. The most fundamental learning model specific to Hamitonian systems is proposed in \cite{greydanus2019hamiltonian} named Hamiltonian neural networks (HNNs), which uses a standard neural network $\widetilde{H}$ to approximate the Hamiltonian $H$ instead of the total vector field $J^{-1}\nabla H$. The input to HNNs are the phase points as well as their derivatives. If only time-dependent discrete phase points are available, a numerical integrator has to be applied to the data to construct the loss. A subsequent work in  \cite{chen2020symplectic} provided the recurrent version of HNNs, namely the symplectic recurrent neural networks (SRNNs). In addition, it experimentally justified that the numerical integrator applied by HNN is preferred to be a symplectic one. Regarding this issue, \cite{zhu2020deep} theoretically proved the necessity of symplectic integration for HNN according to the theory of the inverse modified equation. Both HNNs and SRNNs are, in fact, the indirect methods to identify the flow of the system, by recovering the Hamiltonian $H$ first, then performing prediction using a numerical integrator (better be symplectic) again to solve the learned system. Hence, the HNN-based models are inefficient in the prediction process as well as in the training process, due to the need to compute the gradient of $\widetilde{H}$. Another strategy is to learn the phase flow of the Hamiltonian system directly, based on the prior knowledge of the symplecticity of the Hamiltonian flow as aforementioned. \cite{Chang2017Reverse} used the two-layer Hamiltonian network constructed by the Verlet method \citep{hairer2006geometric} to achieve reversibility and symplecticity. \cite{bondesan2019learning} designed the architecture using the proposed symplectic additive coupling layer as its activation layer, and the pre–Iwasawa decomposed symplectic matrix \citep{de2006symplectic} as its symplectic linear layer. Moreover, \cite{li2020neural} provided a symplectic transformation by employing the real NVP \citep{dinh2016density}. In recent work in \cite{tong2020symplectic} the authors constructed symmetric networks in Taylor expansion form to learn the gradient of the Hamiltonian, then combined them together by a fourth-order symplectic integrator to constitute a symplectic map.

All of the aforementioned symplectic-structured networks lack the theoretical guarantees for their representability, and especially some of them are indeed unable to approximate arbitrary symplectic maps. Additionally, most of them require the learned system to be a separable Hamiltonian system, defined as follows:
\begin{definition}
The Hamiltonian system (\ref{eq:hm_sys}) is separable if $$H(p,q)=T(p)+U(q),\quad p,q\in\R^d.$$
\end{definition}

In this work, we develop symplectic networks (SympNets) to learn the symplectic flow of the Hamiltonian system. In fact, SympNets are able to approximate arbitrary symplectic maps within the set of symplectic maps itself. To the best of our knowledge, this is the first work which can achieve this result with theoretical guarantees. Prior knowledge is incorporated in the sense that the searching space of the neural network is greatly reduced, and the optimization can be performed more effectively. We list below several key advantages of SympNets that will be documented in detail later:
\begin{itemize}
    \item SympNets are able to approximate arbitrary symplectic maps in the $C^r$ norm given appropriate activation functions, such as the sigmoid, hence, they are able to learn the phase flow of arbitrary Hamiltonian systems.
    \item SympNets do not require the learned Hamiltonian systems to be separable.
    \item SympNets can learn the continuous time evolution of dynamics in extended version as stated in Section \ref{subsubsec:learn_irr}.
    \item SympNets can handle the data points resulting from long time steps.
    \item SympNets are highly efficient in training and prediction, as they behave like a standard neural network without the need of extra computation of the gradient during both training and prediction processes or the need of performing numerical integration in the prediction stage as HNN-based models do.
    \item SympNets show great generalization power with an incredibly small network size, as shown in the experiments of the pendulum example, reflecting the expressivity of SympNets.
    \item SympNets are reversible so that the values at the forward passing stage need not be stored.
    \item SympNets can be extended to recurrent version without any modification, compared to SRNNs.
\end{itemize}

The rest of this paper is organized as follows. Section \ref{sec:setup} briefly summarizes the main problem we aim to solve. The detailed process of constructing the SympNets is shown in Section \ref{sec:architecture}. In Section \ref{sec:theory}, we present the theoretical results for SympNets. Section \ref{sec:num_res} presents the simulation results for several Hamiltonian systems. A summary is provided in the last section.

\section{Problem setup} \label{sec:setup}
We apply a neural network model to learn the phase flow of the Hamiltonian system from data. Similar to what numerical integrators do, the trained network is used to compute the phase point after time step $h$ of the start point $y_{0}$, i.e., the input is phase point $y_{0}$ while the output is the phase point $y_{1}=\phi_{h}(y_{0})$.

Assume that the phase flows of (\ref{eq:hm_sys}) are constrained in a compact space $W$. We first choose some phase points from $W$, denoted by $\{x_{i}\}_{1}^{N}$, and then obtain the value of time-$h$ flow $\{y_{i}=\phi_{h}(x_{i})\}_{1}^{N}$ by a high-order symplectic integrator \citep{hairer2006geometric}. Naturally,
\begin{equation*}
    \mathcal{T} = \{(x_{i},y_{i})\}_{1}^{N}
\end{equation*}
is viewed as the training set for learning.
The neural network $\Phi_{h}$ as numerical integrator can be learned by minimizing the
mean-squared-error loss
\begin{equation*}
    MSE = \frac{1}{2d\cdot N}\sum_{i=1}^{N}\|\Phi_{h}(x_{i})-y_{i}\|^{2}.
\end{equation*}
If no prior is placed on $\Phi_h$, it may not possess the property of symplecticity as an integrator, which means that the Hamiltonian may not be conserved in a long-time integration. In other words,  we should carefully design $\Phi_h$ to make sure it is intrinsically symplectic, if we want to make
accurate long term prediction based on the learned model. The architecture of $\Phi_h$ will be shown in the next section.
\section{Architecture} \label{sec:architecture}

\begin{figure*}[htbp]
    \centering
    \includegraphics[width=0.99\textwidth]{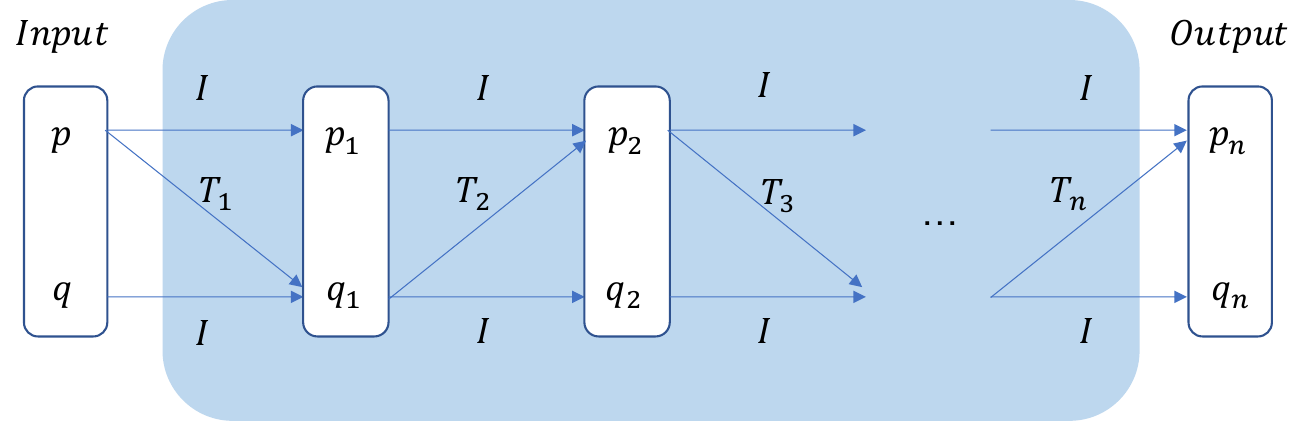}
    \caption{\textbf{Architecture of the SympNets.} The SympNets can be seen as a neural network with the unit triangular connection pattern, which guarantees symplecticity. Here, $T_i$ can be chosen as $S$, $\tilde{\sigma}$ or $\hat{\sigma}$ (defined in Section \ref{sec:architecture}), depending on which type of module it belongs to. Two main types of SympNets, namely LA-SympNets and G-SympNets, are considered in this paper. For LA-SympNets, $T_i$ are chosen to be $S$ or $\tilde{\sigma}$ following a specific order, while for G-SympNets all the $T_i$ are chosen to be $\hat{\sigma}$.}
    \label{fig:architecture}
\end{figure*}

Our architecture design philosophy is based on the fact that the composition of symplectic transformations is again symplectic. In order to construct the destination symplectic map, we make an effort to search for simple linear/nonlinear symplectic maps as the building blocks of the network. We note that the building blocks should be easily parameterized so that they can be efficiently trained. An illustration of the proposed architecture is presented in Fig.~\ref{fig:architecture}.

For convenience, we employ notations used often for matrices and matrix-like maps. In this paper,  $(\cdot)$ denotes a matrix, such as \begin{equation*}
    \begin{pmatrix}A_1 & A_2\\A_3 & A_4\end{pmatrix}\in \R^{2d\times2d},
\end{equation*}
representing the $2d\times2d$-blocked matrix with $A_1,A_2,A_3,A_4\in\R^{d\times d}$, while $[\cdot]$ denotes a matrix-like map, such as
\begin{equation*}
    \begin{bmatrix}f_1 & f_2\\f_3 & f_4\end{bmatrix}:\R^{2d}\to\R^{2d},\quad\begin{bmatrix}f_1 & f_2\\f_3 & f_4\end{bmatrix}\begin{pmatrix}p \\ q\end{pmatrix}:=\begin{pmatrix}f_1(p)+f_2(q)\\f_3(p)+f_4(q)\end{pmatrix},
\end{equation*}
representing the $2d\times2d$-blocked matrix-like map with $f_i:\R^{d}\to\R^{d}$. Sometimes by an abuse of notation, we also represent by the matrix $A\in \R^{d\times d}$ the linear map $p\to Ap$ for $p\in \R^d$. Hence, the identity matrix and the zero matrix $I,0$ may represent the identity map and the zero map, respectively, when they are used in $[\cdot]$.

One of the simplest family of symplectic map from $\R^{2d}$ to $\R^{2d}$ using notations defined above is
\begin{equation}\label{eq:gradV}
    f_{up}\begin{pmatrix}p\\q\end{pmatrix}  = \begin{bmatrix}I & \nabla V\\ 0 & I\end{bmatrix}\begin{pmatrix}p\\q\end{pmatrix},
    \quad
    f_{low}\begin{pmatrix}p\\q\end{pmatrix}  = \begin{bmatrix}I & 0\\ \nabla V & I\end{bmatrix}\begin{pmatrix}p\\q\end{pmatrix},
\end{equation}
where $V:\R^d \to \R$ is an arbitrary function with at least $C^1$ regularity, and $\nabla V:\R^d \to \R^d$ is the gradient of $V$ defined by
\begin{equation*}
    \nabla V(x) = \begin{pmatrix}
    \frac{\partial V}{\partial x_1}, \frac{\partial V}{\partial x_2}, \cdots \frac{\partial V}{\partial x_n}
    \end{pmatrix}^T.
\end{equation*}
In fact, the composition of several $f_{up}$ and $f_{low}$ can approximate any symplectic map, according to \ref{app:proof_appro}. Hence one may directly model $V$ as a neural network to obtain a ``symplectic network''. Nevertheless, this approach requires the computation of the gradient of a network and immediately degenerates to a Hamiltonian neural network discretized by a specific symplectic integrator. SympNets are designed to get rid of the step of calculating the gradients.

\subsection{Linear modules}{\label{sec:Lin}}

In reference to the linear modules, let
\begin{equation} \label{eq:lin_up_low}
\begin{split}
    &\ell_{up}\begin{pmatrix} p \\ q \end{pmatrix}=\begin{pmatrix} I & S \\ 0 & I \end{pmatrix}\begin{pmatrix} p \\ q \end{pmatrix} + b,\quad\ell_{low}\begin{pmatrix} p \\ q \end{pmatrix}=\begin{pmatrix} I & 0 \\ S & I \end{pmatrix}\begin{pmatrix} p \\ q \end{pmatrix} + b,\\
    &b\in\R^{2d}, \quad p, q\in\R^{d},
\end{split}
\end{equation}
where $S\in \R^{d\times d}$ is symmetric. Obviously, $\ell_{up}$ and $\ell_{low}$ are linear and symplectic, however, they are too simple to express a general linear symplectic map. In order to strengthen the expressivity of a linear layer, we compound several $\ell_{up}$ and $\ell_{low}$ alternately as
\begin{equation*}
    \mathcal{L}_{n}^{up}\begin{pmatrix} p \\ q \end{pmatrix}=\begin{pmatrix} I & 0/S_{n} \\ S_{n}/0 & I \end{pmatrix}\cdots\begin{pmatrix} I & 0 \\ S_{2} & I \end{pmatrix}\begin{pmatrix} I & S_{1} \\ 0 & I \end{pmatrix}\begin{pmatrix} p \\ q \end{pmatrix}+b,
\end{equation*}
\begin{equation*}
    \mathcal{L}_{n}^{low}\begin{pmatrix} p \\ q \end{pmatrix}=\begin{pmatrix} I & S_{n}/0 \\ 0/S_{n} & I \end{pmatrix}\cdots\begin{pmatrix} I & S_{2} \\ 0 & I \end{pmatrix}\begin{pmatrix} I & 0 \\ S_{1} & I \end{pmatrix}\begin{pmatrix} p \\ q \end{pmatrix}+b.
\end{equation*}
$\mathcal{L}_{n}^{up}$ and $\mathcal{L}_{n}^{low}$ are referred to as the \textbf{linear modules} in the symplectic network. We would like the linear modules to play a similar role as the linear layers do in a fully-connected neural network. Now a problem is raised naturally, that is, are maps like $\mathcal{L}_{n}$ powerful enough to represent any linear symplectic map? The answer is {\em yes}, and we will present details in Section \ref{sec:theory}. It is noteworthy that  during the prediction process, one may merge the triangular blocks into one matrix in advance for $\mathcal{L}_{n}$ to make predictions faster.
In the following, we will denote the set of the linear modules as:
\begin{equation*}
    \mathcal{M}_L=\{\psi|\psi\ is\ a\ linear\ module\}.
\end{equation*}

 Another issue worth mentioning is parameterization. Since most optimization methods in deep learning focus on unconstrained problems, it is necessary to find a representation for these modules which can be freely parameterized. In fact, the unit triangular symplectic matrices
\begin{equation*}
    \begin{pmatrix} I & S \\ 0 & I \end{pmatrix},\quad\begin{pmatrix} I & 0 \\ S & I \end{pmatrix}\quad (S^{T}=S)
\end{equation*}
can be parameterized as
\begin{equation*}
    \begin{pmatrix} I & A+A^{T} \\ 0 & I \end{pmatrix},\quad\begin{pmatrix} I & 0 \\ A+A^{T} & I \end{pmatrix}
\end{equation*}
where $A$ is a square matrix without any constraint.

\subsection{Activation modules}\label{sec:Act}

We aim to build a simple nonlinear symplectic module, which plays a similar role as the activation layer in a standard fully-connected neural network. The module is designed as
\begin{equation*}
    \begin{pmatrix} P \\ Q \end{pmatrix}=\Phi\begin{pmatrix} p \\ q \end{pmatrix}=\begin{pmatrix} p \\ diag(a)\sigma(p)+q \end{pmatrix},\quad p,q\in\R^d,
\end{equation*}
where $a\in\R^d$ is the parameter to learn, and $\sigma:\R\to\R$ is the activation function acting element-wise as $\sigma(p)=(\sigma(p_{1}),\cdots,\sigma(p_{d}))^{T}$ by a slight abuse of notation. One may readily check that this map is symplectic because it can be written in the form of (\ref{eq:gradV}) with $V(p)=a^T\cdot (\int\sigma)(p)$, where $\int\sigma$ is the antiderivative of $\sigma$. For convenience, we denote this map by
\begin{equation*}
    \mathcal{N}\begin{pmatrix} p \\ q \end{pmatrix}=\begin{bmatrix} I & 0 \\ \Tilde{\sigma}_a & I \end{bmatrix}\begin{pmatrix} p \\ q \end{pmatrix}:=\begin{pmatrix} p \\ diag(a)\sigma(p)+q \end{pmatrix}.
\end{equation*}
 Similar to (\ref{eq:lin_up_low}), we specifically define
\begin{equation*}
    \mathcal{N}_{up}\begin{pmatrix} p \\ q \end{pmatrix}=\begin{bmatrix} I & \Tilde{\sigma}_a \\ 0 & I \end{bmatrix}\begin{pmatrix} p \\ q \end{pmatrix},\quad\mathcal{N}_{low}\begin{pmatrix} p \\ q \end{pmatrix}=\begin{bmatrix} I & 0 \\ \Tilde{\sigma}_a & I \end{bmatrix}\begin{pmatrix} p \\ q \end{pmatrix}.
\end{equation*}
$\mathcal{N}_{up}$ and $\mathcal{N}_{low}$ are referred to as the \textbf{activation modules} of a symplectic network. This layer plays the same role as activation layer in a standard fully-connected neural network. The universal approximation theorem of neural networks states that any continuous function can be approximated by the composition of linear units and activation units under certain constraints. Similarly, it will be shown in section \ref{sec:theory} that any symplectic map can be approximated by a composition of linear modules and activation modules.
In the following, we will denote the set of the activation modules as:
\begin{equation*}
    \mathcal{M}_A=\{\psi|\psi\ is\ an\ activation\ module\}.
\end{equation*}

\subsection{Gradient modules}
In addition to the modules provided in section \ref{sec:Lin} and \ref{sec:Act}, we offer an alternative choice, called the gradient module. This module will not change the approximation properties of the network, however, it offers an option, which may converge faster and result in lower testing error in some cases.

Let us define a symplectic map given an activation function $\sigma$ in the following way:
\begin{equation*}
    \mathcal{G}\begin{pmatrix} p \\ q \end{pmatrix}=\begin{bmatrix} I & 0 \\ \hat{\sigma}_{K,a,b} & I \end{bmatrix}\begin{pmatrix} p \\ q \end{pmatrix}:=\begin{pmatrix} p \\ K^Tdiag(a)\sigma(Kp+b)+q \end{pmatrix},
\end{equation*}
where $b\in \mathbb{R}^{n}$, $K\in \mathbb{R}^{n\times d}$, $a\in \mathbb{R}^{n}$, and $n$ is a positive integer regarded as the width of the module. In practice, we let $n>d$ to increase the expressivity of the module. To see that $\mathcal{G}$ is indeed symplectic, one only needs to check it with (\ref{eq:gradV}), details are omitted here. Now we define
\begin{equation*}
    \mathcal{G}_{up}\begin{pmatrix} p \\ q \end{pmatrix}=\begin{bmatrix} I & \hat{\sigma}_{K,a,b} \\ 0 & I \end{bmatrix}\begin{pmatrix} p \\ q \end{pmatrix},\quad\mathcal{G}_{low}\begin{pmatrix} p \\ q \end{pmatrix}=\begin{bmatrix} I & 0 \\ \hat{\sigma}_{K,a,b} & I \end{bmatrix}\begin{pmatrix} p \\ q \end{pmatrix}.
\end{equation*}
$\mathcal{G}_{up}$ and $\mathcal{G}_{low}$ are referred to as the  \textbf{gradient modules} of the symplectic network. The ``gradient module'' is named following the principle that $\hat{\sigma}_{K,a,b}$ can approximate an arbitrary $\nabla V$ as shown in \ref{app:proof_appro}. These modules are inspired by the two-layer Hamiltonian network in \cite{Chang2017Reverse}, the symmetric layer in \cite{Ruthotto2018RCNN} and the symplectic additive coupling layer in \cite{bondesan2019learning}.
In the following, we will denote the set of gradient modules as:
\begin{equation*}
    \mathcal{M}_G=\{\psi|\psi\ is\ a\ gradient\ module\}.
\end{equation*}

\subsection{SympNets}
The \textbf{symplectic networks (SympNets)} can be informally defined as the composition of linear, activation and gradient modules. More formally, we have the following definition:

\begin{definition}
Consider $\{v_{i}\}_{1}^{k}\subset \mathcal{M}_L\cup \mathcal{M}_A \cup \mathcal{M}_G$, where $\mathcal{M}_L$, $\mathcal{M}_A$ and $\mathcal{M}_G$ are the set of linear, activation and gradient modules respectively. Let
\begin{equation*}
    \psi=v_{k}\circ v_{k-1}\circ\cdots\circ v_{1}.
\end{equation*}
Any such $\psi$ is called \textbf{symplectic network (SympNet)}. Furthermore, we define the collection of SympNets as
\begin{equation*}
    \Psi=\{\psi|\psi\ is\ a\ SympNet\}.
\end{equation*}
\end{definition}
Theoretically, the SympNets enjoy great algebraic and approximation properties, which will be discussed in the next section. Practically, a SympNet is highly flexible in the sense that different modules can be assembled in many different ways. The users could apply a neural architecture search (NAS) algorithm to find out the best way to assemble these modules. Here, we introduce two easily realizable ways of formulating a symplectic network, for the purpose of both proving theorems and performing numerical simulations.
\begin{definition}
Consider $\{v_{i}\}_{1}^{k+1} \subset \mathcal{M}_L$, $\{w_{i}\}_{1}^{k}\subset \mathcal{M}_A$. Let
\begin{equation*}
    \psi=v_{k+1}\circ w_{k}\circ v_{k}\circ\cdots\circ w_{1}\circ v_{1},
\end{equation*}
where $\psi$ is called \textbf{LA-SympNet}. We define the collection of LA-SympNets as
\begin{equation*}
    \Psi_{LA}=\{\psi|\psi\ is\ a\ LA\text{-}SympNet\}.
\end{equation*}
\end{definition}
\begin{definition}
Consider $\{u_{i}\}_{1}^{k}\subset \mathcal{M}_G$. Let
\begin{equation*}
    \psi=u_{k}\circ u_{k-1}\circ\cdots\circ u_{1},
\end{equation*}
where $\psi$ is called \textbf{G-SympNet}. We define the collection of G-SympNets as
\begin{equation*}
    \Psi_{G}=\{\psi|\psi\ is\ a\ G\text{-}SympNet\}.
\end{equation*}
\end{definition}
Note that both $\Psi_{LA}, \Psi_{G} \subset \Psi$. $\Psi_{LA}$ can be considered as the alternated composition of linear and activation modules while $\Psi_{G}$ can be thought of as the simple combination of gradient modules. We will show that both $\Psi_{LA}$ and $\Psi_{G}$ are dense in the set of all the symplectic maps given an appropriate activation function in section \ref{sec:theory}.

\section{Theory of SympNets} \label{sec:theory}
\subsection{Algebraic properties}
\begin{theorem}[Algebraic structure]\label{thm:alg_str}
The collection of all the SympNets $\Psi$ is a group in the sense of map composition.
\end{theorem}
\begin{proof}
We know that the identity map $I \in \mathcal{M}_L\subset \Psi$ is the identity element of $\Psi$ (group ``multiplication'' is given by map composition). Moreover, the associative law and the closure obviously hold by the definition of $\Psi$. What we need to confirm is that there exists an inverse element for any $\psi$, i.e., $\psi^{-1}\in \Psi$. Observe that
\begin{equation*}
\begin{bmatrix} I & f \\ 0 & I \end{bmatrix}^{-1}=\begin{bmatrix} I & -f \\ 0 & I \end{bmatrix},\quad \begin{bmatrix} I & 0 \\ f & I \end{bmatrix}^{-1}=\begin{bmatrix} I & 0 \\ -f & I \end{bmatrix},
\end{equation*}
where $f:\mathbb{R}^d \to \mathbb{R}^d $. By substituting $f = S$, $\tilde{\sigma}_a$ and $\hat{\sigma}_{K,a,b}$ respectively,
we derive that for any
     $\mathcal{L} \in \mathcal{M}_L,\ \mathcal{N} \in \mathcal{M}_A,\ \mathcal{G} \in \mathcal{M}_G$,
    it holds that
    \begin{equation*}
        \mathcal{L}^{-1} \in \mathcal{M}_L \subset \Psi,\ \mathcal{N}^{-1} \in \mathcal{M}_A \subset \Psi,\ \mathcal{G}^{-1} \in \mathcal{M}_G \subset \Psi.
    \end{equation*}
Now we consider an arbitrary SympNet $\psi=v_{k}\circ v_{k-1}\circ\cdots\circ v_{1}\in \Psi.$
It can be seen that
\begin{equation*}
\psi^{-1}=v_{1}^{-1}\circ\cdots\circ v_{k-1}^{-1}\circ v_{k}^{-1}\in \Psi.
\end{equation*}
We therefore conclude that $\Psi$ is a group.
\end{proof}
Being a group endows $\Psi$ with many practically useful properties. One direct implication of theorem \ref{thm:alg_str} is the following:
\begin{corollary}
Any SympNet $\psi \in \Psi$ is reversible.
\end{corollary}
Reversibility means that there is an analytic inverse and the value of the neural network at each layer can be computed from the output of the entire network, i.e., once $\psi(x)$ is known, we can obtain the value of $v_{i}\circ\cdots \circ v_1(x)$ for each $1\leq i\leq k$. This implies that these values are unnecessary to be stored at the forward passing stage, since they can be computed directly at the backward propagation stage, which enables a memory-efficient way of implementing the neural network. This type of neural network has applications in image classification and generative modeling \citep{dinh2014nice, dinh2016density, Chang2017Reverse, pmlr-v97-behrmann19a}. Ideas in this direction can be further explored in the future.
\begin{example}
Here is an example for the reverse SympNet. If
\begin{equation*}
\begin{split}
    \begin{pmatrix}P \\ Q\end{pmatrix}=&\psi\begin{pmatrix}p \\ q\end{pmatrix} \\ =&\begin{pmatrix}I & S_3\\0 & I\end{pmatrix}\begin{bmatrix}I & diag(a)\sigma\\0 & I\end{bmatrix}\begin{pmatrix}I & 0\\S_2 & I\end{pmatrix}\begin{pmatrix}I & S_1\\0 & I\end{pmatrix}\begin{pmatrix}p \\ q\end{pmatrix},
\end{split}
\end{equation*}
then
\begin{equation*}
\begin{split}
    \begin{pmatrix}p \\ q\end{pmatrix} =&\psi^{-1}\begin{pmatrix}P \\ Q\end{pmatrix} \\
    =&\begin{pmatrix}I & -S_1\\0 & I\end{pmatrix}\begin{pmatrix}I & 0\\-S_2 & I\end{pmatrix}\begin{bmatrix}I & -diag(a)\sigma\\0 & I\end{bmatrix}\begin{pmatrix}I & -S_3\\0 & I\end{pmatrix}\begin{pmatrix}P \\ Q\end{pmatrix}.
\end{split}
\end{equation*}
\end{example}
\begin{theorem}\label{thm:alg2}
The collection of LA(G)-SympNets $\Psi_{LA}$($\Psi_{G}$) is a group.
\end{theorem}
\begin{proof}
It is similar as the proof of Theorem \ref{thm:alg_str}.
\end{proof}
\subsection{Approximation properties}
In this section, we wil present three main theorems regarding approximation properties of the SympNets. We start by introducing a few notations, which will be used later.
Denote the set of symplectic matrices as
\begin{equation*}
SP=\{H\in\R^{2d\times 2d}|H^{T}JH=J\}.
\end{equation*}
 Similarly, we denote the set of $C^{r}$ smooth symplectic map on an open set $U\subset\R^{2d}$ as
\begin{equation*}
    \mathcal{SP}^r(U) = \left\{\Phi\in C^r(U;\R^{2d})\Bigg|\left(\frac{\partial \Phi}{\partial x}\right)^T J \left(\frac{\partial \Phi}{\partial x}\right) = J \right\},\quad r\geq 1.
\end{equation*}
Also, denote

\begin{equation*}
\begin{split}
    L_{n} = \Bigg\{& \begin{pmatrix} I & 0/S_{n}\\ S_{n}/0 & I \end{pmatrix}\cdots\begin{pmatrix} I & 0 \\ S_{2} & I \end{pmatrix}\begin{pmatrix} I & S_{1} \\ 0 & I \end{pmatrix}\Bigg| \\
    & S_{i}\in\R^{d\times d},S_{i}^{T}=S_{i},i=1,2,\cdots,n \Bigg\}
\end{split}
\end{equation*}
where the unit upper triangular symplectic matrices and the unit lower triangular symplectic matrices appear alternately. It is clear that $L_{m}\subset L_{n}\subset SP$ for all integers $1\leq m\leq n$.
\begin{theorem}\label{thm:repre_thm}
$SP=L_9$. Thus, $\mathcal{M}_L$ consists of all the linear symplectic maps.
\end{theorem}
\begin{proof}
It is known from our previous work \citep{jin2019unit} that $SP=L_9$.
\end{proof}
The above theorem indicates that the linear modules can parameterize any linear symplectic map. Moreover, the depth of each linear module need not be larger than 9. In \cite{jin2019unit}, we systematically present several existing modern factorizations of the matrix symplecic group, and propose the unit triangular factorization described as Theorem \ref{thm:repre_thm}. This factorization induces the unconstrained parametrization of the matrix symplectic group by replacing the block $S_{i}$ with $A_{i}+A_{i}^{T}$. It enables us to make use of the symplectic matrix as a module in a deep neural network, just like what we are doing here.

Restrictions on activation functions have to be made before any type of approximation theorem of neural networks can be given. Here, we introduce some necessary notations first. Let $D^\alpha$ be the differential operator, where $\alpha=(\alpha_1,\cdots,\alpha_m)$ with non-negative integers $\alpha_i$ is an ordered set of differential indexes. As an example, for $f\in C^{\infty}(\R^m)$, we have
\begin{equation*}
    D^\alpha f=\frac{\partial^{|\alpha|}f}{\partial {x_1}^{\alpha_1}\cdots\partial {x_m}^{\alpha_m}},\quad |\alpha|=\alpha_1+\cdots+\alpha_m.
\end{equation*}
Furthermore, we define the norm on $C^r(W;\R^{n})$ as
\begin{equation*}
\begin{split}
    &\norm{f}_{C^r(W;\R^{n})}=\sum_{|\alpha|\leq r}\max_{1\leq i\leq n}\sup_{x\in W} |D^\alpha f_i(x)|,\\ &f=(f_1,\cdots,f_{n})^T\in C^r(W;\R^{n}),
\end{split}
\end{equation*}
for a compact set $W\subset \R^{m}$.

\begin{definition}
Let $r\in\{0\}\cup\mathbb{N}^*$ be given. $\sigma$ is $r$-finite if $\sigma\in C^r(\mathbb{R})$ and $0<\int|D^r(\sigma)|d\lambda <\infty$. Here $\mathbb{N}^*$ is the set of positive integers and $\lambda$ is the Lebesgue measure on $\mathbb{R}$.
\end{definition}
One of the most commonly used activation functions, the sigmoid function, satisfies this condition for any $r\in \mathbb{N}^*$. We will formalize and show this result in lemma \ref{lem:sig}.

\begin{definition}
Let $m,n\in \mathbb{N}^*,r\in\{0\}\cup\mathbb{N}^*$ be given, $U\subset \R^{m}$ is an open set, $S_1 \subset C^r(U;\R^{n})$, then we say $S_2$ is $r$-uniformly dense on compacta in $S_1$ if $S_2\subset S_1$ and for any $f\in S_1$, compact $W\subset U$ and any $\epsilon>0$, there exists $g\in S_2$ such that $\norm{f - g}_{C^r(W;\R^{n})}<\epsilon$.

\end{definition}

With the above concepts, next we present the {\em universal approximation theorems}  for SympNets, and provide their  proofs in \ref{app:proof_appro}.

\begin{theorem}[Approximation theorem for LA-SympNets]\label{thm:symp_LA}

For any $r\in\mathbb{N}^*$ and open $U\subset \R^{2d}$, the set of LA-SympNets $\Psi_{LA}$ is $r$-uniformly dense on compacta in $\mathcal{SP}^r(U)$ if the activation function $\sigma$ is $r$-finite.
\end{theorem}

\begin{theorem}[Approximation theorem for G-SympNets]\label{thm:symp_G}

For any $r\in\mathbb{N}^*$ and open $U\subset \R^{2d}$, the set of G-SympNets $\Psi_G$ is $r$-uniformly dense on compacta in $\mathcal{SP}^r(U)$ if the activation function $\sigma$ is $r$-finite.

\end{theorem}

Following Theorems \ref{thm:symp_LA} and \ref{thm:symp_G}, $\Psi$ is also $r$-uniformly dense on compacta in $\mathcal{SP}^r(U)$. Moreover, since $\Psi$ and $\mathcal{SP}^r(\R^{2d})$ are groups, $\Psi\subset \mathcal{SP}^r(\R^{2d})$, we have that $\Psi$ is a dense subgroup of $\mathcal{SP}^r(\R^{2d})$.

Theorems \ref{thm:symp_LA} and \ref{thm:symp_G} give the general criterion for a SympNet to possess the universal approximation property. It is worth mentioning that the sigmoid function satisfies the condition.
\begin{lemma}\label{lem:sig}
    The sigmoid activation, $\sigma(x) = \frac{1}{1+e^{-x}}$, is $r$-finite for any $r\in \mathbb{N}^*$.
\end{lemma}
\begin{proof}
$\sigma'(x) = \frac{e^{-x}}{(1+e^{-x})^2}>0$, so $\int |\sigma'(x)| d\lambda = \int \sigma'(x) d\lambda= 1$.
By mathematical induction, one can show that when
$n\geq2$,
\begin{equation*}
    \sigma^{(n)}(x) = \sigma'(x)P^{(n-1)}(\sigma(x)),
\end{equation*}
where $P^{(n-1)}(\cdot)$ is an $(n-1)$-th order polynomial, then
\begin{equation*}
\begin{split}
    0<\int|\sigma^{(r)}(x)|d\lambda \leq& \int|\sigma'(x)|d\lambda \cdot(\sup_{x\in\R}|P^{(r-1)}(\sigma(x))|) \\
    = &\sup_{y\in[0,1]}|P^{(r-1)}(y)| \\
    < &\infty.
\end{split}
\end{equation*}
Therefore $\sigma$ is $r$-finite for any $r\in \mathbb{N}^*$.
\end{proof}
\begin{corollary}
The set of sigmoid-activated LA(G)-SympNets is $r$-uniformly dense on compacta in $\mathcal{SP}^r(U)$ for any $r\in \mathbb{N}^*$ and open $U\subset\R^{2d}$.
\end{corollary}
Therefore, we will use the sigmoid activation function for all of our simulation experiments
presented below.

\begin{figure*}[htbp]
    \centering
    \includegraphics[width=0.99\textwidth]{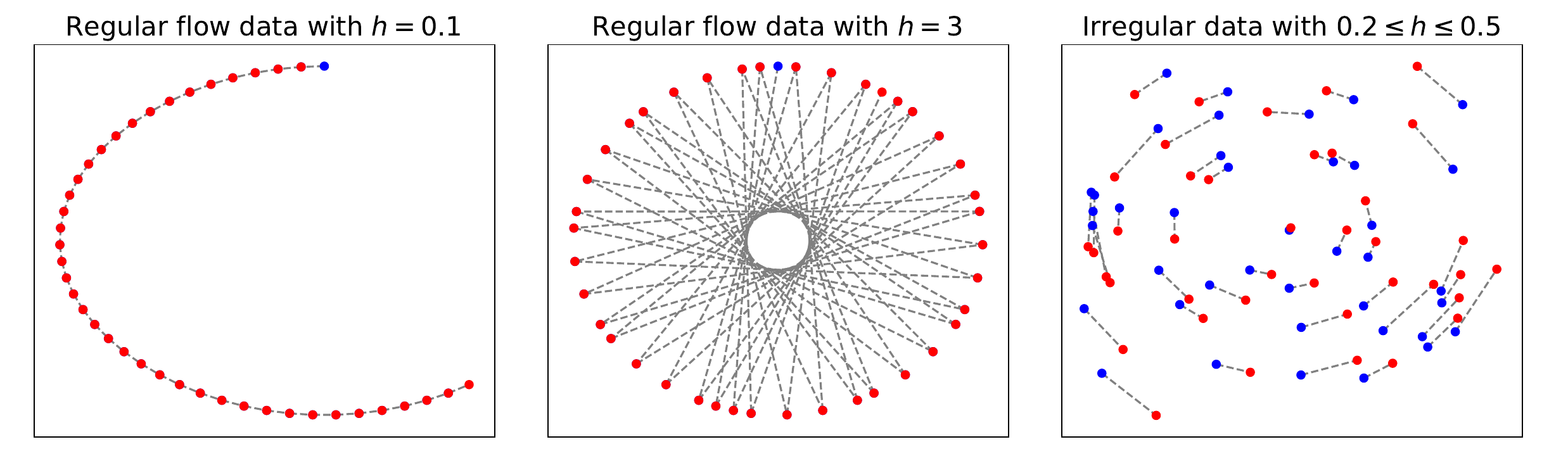}
    \caption{\textbf{Illustrations for datasets of pendulum.} Three types of datasets are used in the experiments of pendulum. A dash line connects a blue dot representing the initial state and a red dot representing the next state after a time step $h$.}
    \label{fig:pend_data}
\end{figure*}

\section{Simulation results} \label{sec:num_res}
Besides their universal approximation power, SympNets, specifically LA/G-SympNets, possess many other desirable properties in the sense that: first, they are able to generalize using limited amount of data with a small parameter space. Second, they can handle long time step prediction based models. Third, they can handle irregularly sampled data. Fourth, they can deal with non-separable Hamiltonian. Last but not least, they scale well in high dimensions. We illustrate these properties of SympNets by solving three different numerical prediction tasks, namely the pendulum, double pendulum and three-body problems. The codes are published in GitHub (https://github.com/jpzxshi/sympnets).

The benchmark method used for comparison in this section is HNN~\citep{greydanus2019hamiltonian}. The main objective to be minimized in HNN is
\begin{equation*}
    \left\|\frac{dy}{dt}-J^{-1}\nabla \widetilde{H}(y)\right\|,
\end{equation*}
where $y\in\R^{2d}$, $\widetilde{H}$ is a standard neural network. In many application scenarios, the derivative of vector fields $\frac{dy}{dt}$ is unknown, so it should be approximated using numerical discretization integrators. In fact, symplectic integrators should be applied, as is numerically justified in \cite{chen2020symplectic} and theoretically proved in \cite{zhu2020deep}. In all of our experiments, we use the midpoint rule, a symplectic integrator of order 2, to approximate the objective:
\begin{equation*}
    \left\|\frac{x_{i+1}-x_{i}}{h}-J^{-1}\nabla \widetilde{H}(\frac{x_{i}+x_{i+1}}{2})\right\|.
\end{equation*}
Once $\Tilde{H}$ has been learned, we perform prediction using a 4th order symplectic integrator with finer time step to ensure the correctness of the predictions of HNNs, which indeed costs much more time compared to SympNets that make predictions directly. Since symplectic methods are applied in both training and testing procedures, we will refer to the enhanced baseline model as S-HNNs, where S stands for symplectic.
In this paper, we do not require the Hamiltonians to be separable a priori for any of the test cases, so multistep or recurrent training in \cite{chen2020symplectic} will not be considered for S-HNNs, especially for the case of the double pendulum that is indeed non-separable.
\begin{table*}[htbp]
    \centering
    \begin{tabular}{c|c|c|c|c|c}
        \toprule
        Problem & Type & Depth & Sublayers & Width & Parameters \\
        \midrule
         & S-HNN & 4 & N/A & 30 & 2K  \\
        Pendulum: flow data & LA-SympNet & 3 & 2 & N/A & 14 \\
         & G-SympNet &5 & N/A & 30 & 0.5K \\
        \midrule
         & S-HNN & 4 & N/A & 30 & 2K  \\
        Pendulum: irregular data & LA-SympNet & 5 & 4 & N/A & 34 \\
         & G-SympNet &5 & N/A & 30 & 0.5K \\
         \midrule
         & S-HNN & 4 & N/A & 50 & 5K \\
        Double pendulum & LA-SympNet & 8 & 5 & N/A & 0.2K \\
         & G-SympNet & 8 & N/A & 50 & 2K \\
         \midrule
         & S-HNN & 6 & N/A & 50 & 11K \\
         Three-body & LA-SympNet & 20 & 4 & N/A & 3K \\
         & G-SympNet &20 & N/A & 50 & 8K \\
        \bottomrule
    \end{tabular}
    \caption{\textbf{Model architecture.} S-HNN uses fully-connnected neural network (FNN) as its approximator to the Hamiltonian. Depth represents the number of linear layers (linear modules) used in S-HNN (LA-SympNet), while for G-SympNet it is equal to the number of gradient modules. The number of sublayers for LA-SympNet is the number of $\ell_{up}$ or $\ell_{low}$ used to constitute each linear module. The width for G-SympNet is $n$, the row dimension of $K$ in the definition of gradient module.}
    \label{tab:model_params}
\end{table*}
\begin{table*}[htbp]
    \centering
    \begin{tabular}{c|c|c|c}
        \toprule
        Problem & Type & Learning rate & Epochs \\
        \midrule
         & S-HNN & 0.001 & 100000 \\
        Pendulum: flow data & LA-SympNet & 0.001 & 100000 \\
         & G-SympNet & 0.001 & 100000 \\
        \midrule
         & S-HNN & 0.001 & 100000 \\
        Pendulum: irregular data & LA-SympNet & 0.01 & 100000 \\
         & G-SympNet & 0.01 & 100000 \\
         \midrule
         & S-HNN & 0.001 & 300000 \\
        Double pendulum & LA-SympNet & 0.001 & 300000 \\
         & G-SympNet & 0.001 & 300000 \\
         \midrule
         & S-HNN & 0.001 & 300000 \\
         Three-body & LA-SympNet & 0.001 & 300000 \\
         & G-SympNet & 0.001 & 300000 \\
        \bottomrule
    \end{tabular}
    \caption{\textbf{Training parameters.} The optimizer is set to be Adam \citep{adam2015} for all the cases. The double pendulum and three-body problems require more epochs to converge than the pendulum problem since those problems are in higher dimensions.}
    \label{tab:training_params}
\end{table*}
\begin{table*}
\centering
\begin{tabular}{|c|c|c|c|c|}
\hline
\multicolumn{2}{|c|}{Type} & LA-SympNet & G-SympNet & S-HNN \\ \hline
\multirow{2}{*}{$h=0.1$} & Test MSE ($\log_{10}$) & $-7.3\pm 0.1$ & $-3.0\pm 0.2$ & $-1.6\pm 0.6$ \\ \cline{2-5}
 & VPT ($\log_{10}$) & $3.5\pm 0.3$ & $1.2\pm 0.1$ & $0.3\pm 0.3$ \\ \hline
\multirow{2}{*}{$h=3$} & Test MSE ($\log_{10}$) & $-6.7\pm 0.4$ & $-5.2\pm 0.5$ & N/A \\ \cline{2-5}
 & VPT ($\log_{10}$) & $4.5\pm 0.4$ & $3.7\pm 0.2$ & N/A \\ \hline
\multirow{2}{*}{Irregular} & Test MSE ($\log_{10}$) & $-4.4\pm 0.4$ & $-4.1\pm 0.5$ & $-3.2\pm 0.2$ \\ \cline{2-5}
 & VPT ($\log_{10}$) & $2.1\pm 0.6$ & $1.8\pm 0.4$ & $1.2\pm 0.1$ \\ \hline
\end{tabular}
\caption{\textbf{Quantitative results for the pendulum.} The test MSE and the VPT are recorded in the form of mean $\pm$ standard deviation in log scale based on 10 independent experiments. SympNets outperform S-HNNs in all test cases.}
\label{tab:pend_table}
\end{table*}
\begin{figure*}[htbp]
    \centering
    \includegraphics[width=0.99\textwidth]{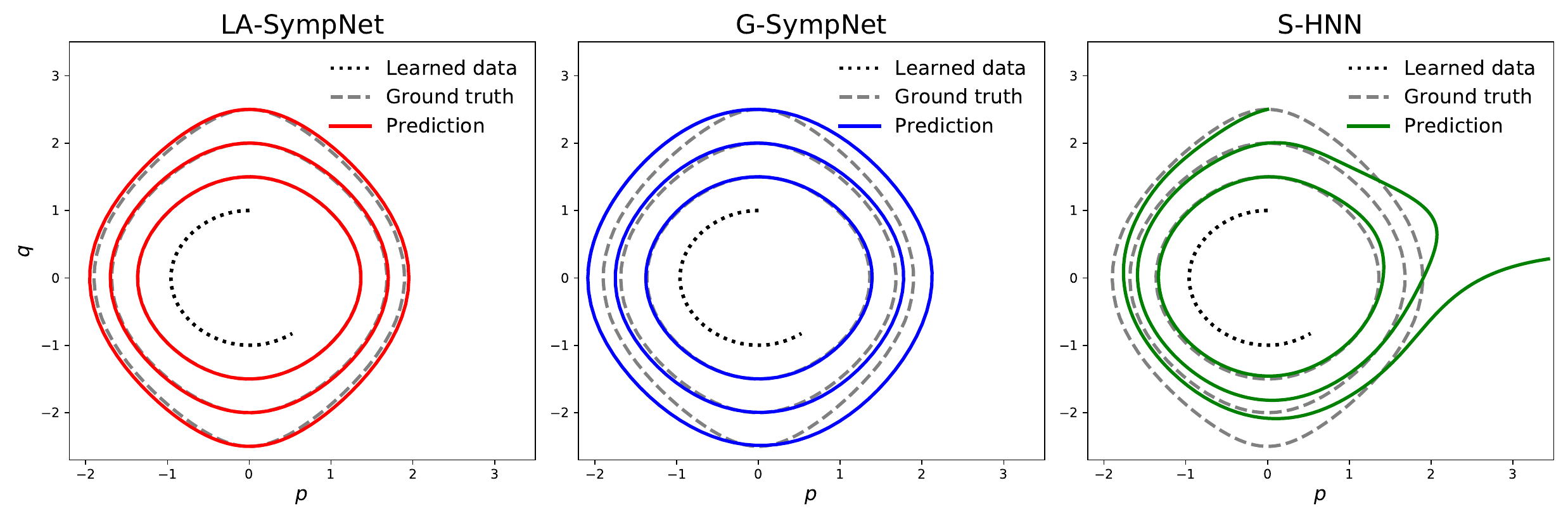}
    \caption{\textbf{Inferences of the outer trajectories starting at $\mathbf{(p,q) = (0, 1.5),(0, 2),(0, 2.5)}$ for $\mathbf{h=0.1}$.} This figure examines the extrapolation power of three different models given training data on a single trajectory starting at $(p,q) = (0, 1)$.}
    \label{fig:pend_outer}
\end{figure*}
\subsection{Hyper-parameters}
Table \ref{tab:model_params} shows the architecture of the models we used for each problem. We see that LA-SympNets require a significantly smaller parameter space, especially in the case of the pendulum with $h=0.1$, where it takes only 14 parameters to achieve the best performance. The activation function is chosen to be sigmoid for SympNets and hyperbolic tangent ($\tanh$) for S-HNNs. We use the normal distribution $\mathcal{N}(0,0.01)$ to initialize each entry of the weight matrices in SympNets, while for S-HNNs, principal orthogonal initialization is applied. The training parameters are presented in Table~\ref{tab:training_params}.

\subsection{The Pendulum problem}
\subsubsection{Datasets and evaluation metric}
Consider the pendulum system with the Hamiltonian
\begin{equation*}
    H(p,q)=\frac{1}{2}p^{2}-\cos(q).
\end{equation*}
In particular, we use three different datasets: (i) flow data with $h = 0.1$, (ii) flow data with $h=3$, (iii) irregular data, to illustrate the first of three properties of SympNets described at the beginning of this section. The detailed definitions of these datasets are shown below and also in Fig.~\ref{fig:pend_data}.

\noindent\textbf{Flow data.} The training dataset consists of $N = 40$ data points on a single trajectory starting from $x_0=(0,1)$ with shared time step $h$. These data points are grouped in pairs before being fed into the neural network,  denoted as $\mathcal{T}=\{(x_{i-1},x_{i})\}_{1}^{N}$, where $x_{i}=\phi_{h}(x_{i-1}),i=1,2,\cdots,N$. The test dataset is given by the $k = 100$ data points following the last point in the training dataset, denoted by $X = (x_{N+1},\cdots,x_{N+k} )$.
After training on $\mathcal{T}$, we use the trained network $\Phi_{h}$ to compute the flow starting at $x_N$ for 100 steps, denoted by $\tilde{X} = (\Tilde{x}_{N+1},\cdots,\Tilde{x}_{N+k} )$. The mean squared error between $X$ and $\tilde{X}$ is taken as the test MSE.

\noindent\textbf{Irregular data.} The training dataset consists of $N = 40$ grouped pairs of points randomly sampled from $[-\sqrt{2},\sqrt{2}]\times[-\frac{1}{2}\pi,\frac{1}{2}\pi]$ with time steps $\{h_i\}_1^N$ randomly chosen in $[0.2, 0.5]$, denoted as $\mathcal{T}=\{([x_i,h_i],y_i)\}_{1}^{N}$, where $y_i=\phi_{h_i}(x_{i}),i=1,2,\cdots,N$. The test dataset is generated by $k=100$ data points on a single trajectory following $x_0=(0, 1)$ with shared time step $h=0.1$, denoted by $X = (x_{1},\cdots, x_{k} )$. Same as in flow data, we generate $\tilde{X} = (\Tilde{x}_{1},\cdots,\Tilde{x}_{k} )$ by the trained network and compute the mean squared error between $X$ and $\Tilde{X}$ as the test MSE. Note that we will explain how to apply the data with additional input $h$ to SympNets later.

We compute the valid prediction time $T_\epsilon$ following a similar definition as in \cite{VLACHAS2020191} in order to evaluate the predictive performance of different models. Suppose we are given the ground truth dataset $x$ and prediction $\tilde{x}$. Let the normalized root mean square error be
\begin{equation*}
    \mathcal{E} (\tilde{x}) = \sqrt{\langle\frac{(x-\tilde{x})^2}{s^2}\rangle},
\end{equation*}
where $s\in \R^{2d}$ is the standard deviation in time of each state component of $x$, and $<\cdot>$ represents spatial average. The valid prediction time of the model is given by
\begin{equation*}
    T_\epsilon = \argmax_{t_f} \{t_f|\mathcal{E}(\tilde{x}(t)) \leq \epsilon, \forall t\leq t_f\}.
\end{equation*}
In other words, $T_\epsilon$ characterizes the longest prediction window that the model remains valid. In this section, $\epsilon$ is set to 0.1.
\begin{figure*}[htbp]
    \centering
    \includegraphics[width=0.99\textwidth]{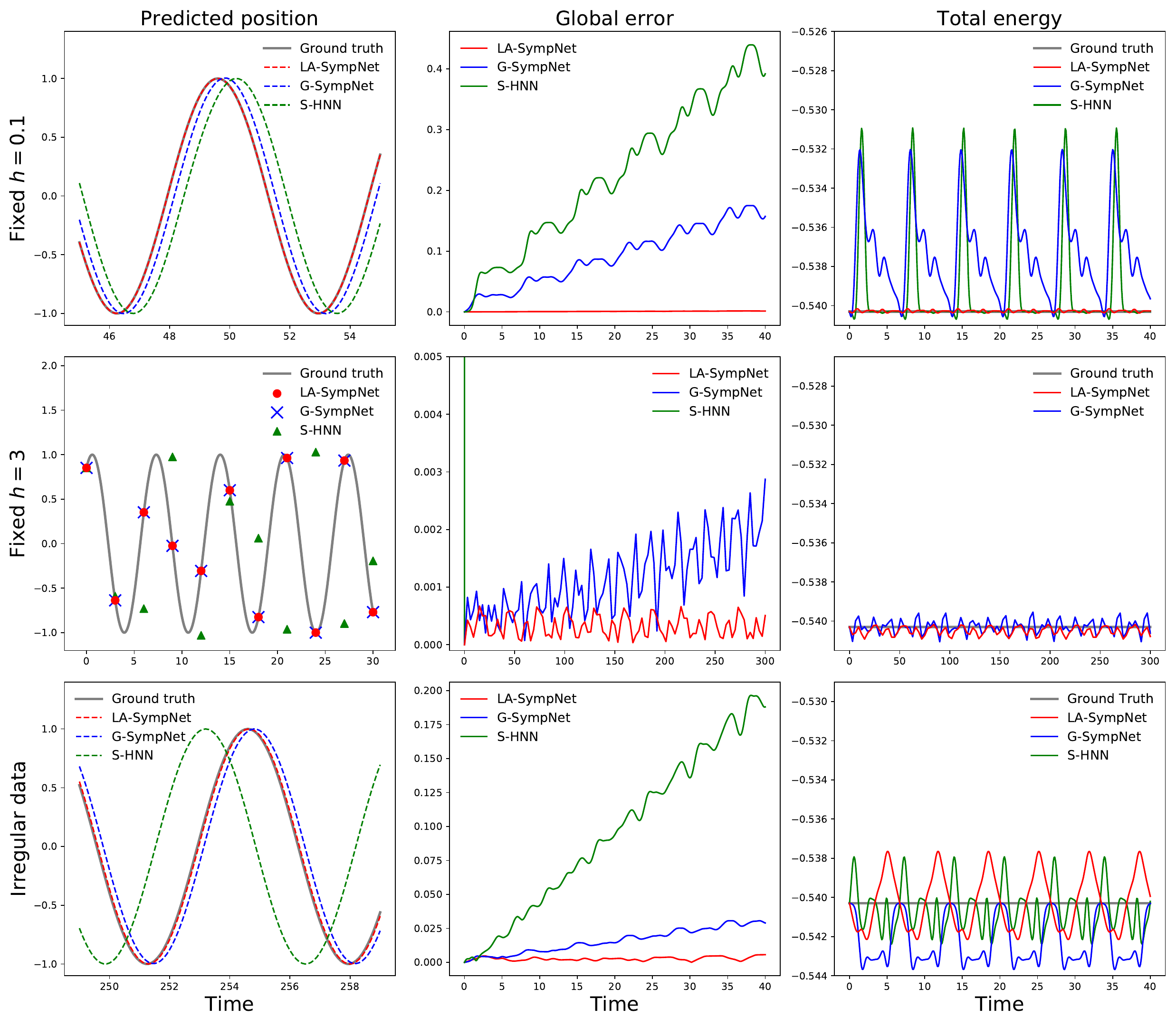}
    \caption{\textbf{Results of LA-SympNets, G-SympNets and S-HNNs on the three datasets for the pendulum system.} (\textbf{Left column}) The predicted positions $q$ for the three datasets. The time windows are chosen so that the differences among the predictions made by the three methods appear. (\textbf{Middle column}) The global errors for the three datasets. It is observed that $Error_{LA}<Error_{G}<Error_{H}$ on all the datasets. (\textbf{Right column}) The total energies for the three datasets. The y-axes of the three subplots are of the same length scale so that the energy fluctuation levels can be clearly seen. The energy of S-HNN for $h=3$ is not shown since it explodes. (\textbf{Total figure})
    LA-SympNets and G-SympNets outperform S-HNNs in all cases.  In fact, LA-SympNets always have the lowest global error. S-HNNs fail to learn the data with large time step due to the time discretization. LA, G and S-HNNs can all preserve the energy correctly except for S-HNN when $h=3$.}
    \label{fig:pend_performance}
\end{figure*}

 \begin{figure*}[htbp]
    \centering
    \includegraphics[width=0.99\textwidth]{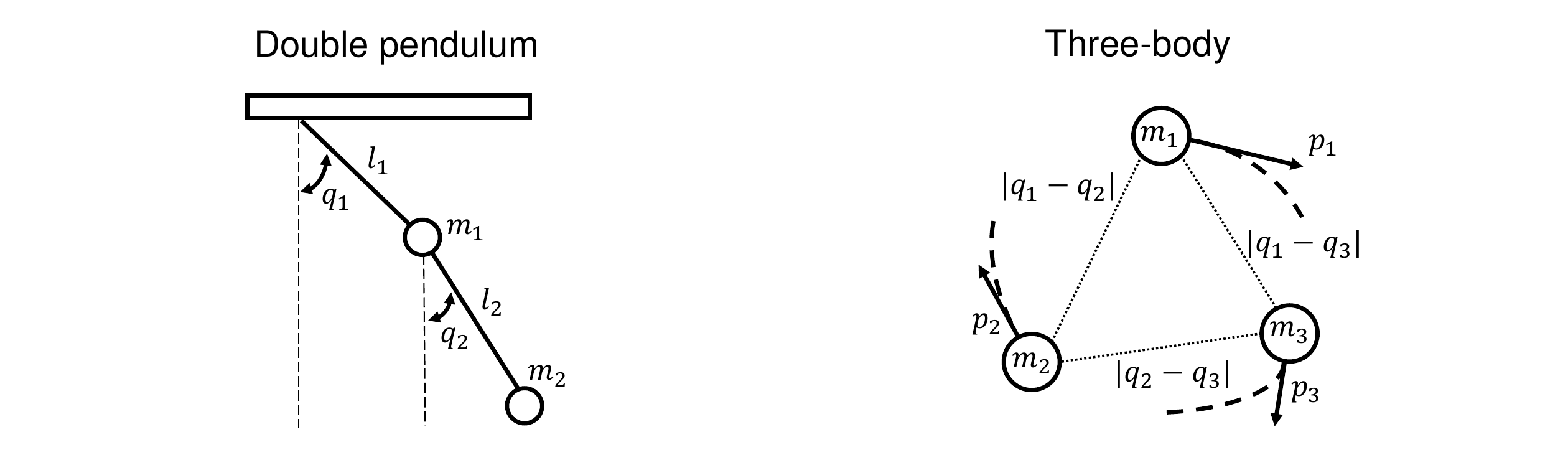}
    \caption{\textbf{Illustrations for the double pendulum and the three-body.} (\textbf{Left}) The double pendulum system consists of a pendulum attached directly to another one. The $i$-th pendulum is made of a ball of mass $m_i$ connected to a massless rigid rod of length $l_i$. (\textbf{Right}) The three-body system consists of three planets of mass $m_i$, the motion of which is governed purely by gravitational force.}
    \label{fig:illus_double_3b}
\end{figure*}

\subsubsection{Learning flows with fixed time steps}
 This problem is more difficult than the pendulum prediction problem in \cite{greydanus2019hamiltonian} in the sense that the training data points do not cover an entire period of the trajectory, as can be seen from Fig.~\ref{fig:pend_data}. Indeed, the problem is ill-posed because there might be more than one Hamilton's equations whose solution could match these data points. Therefore, the learned models are expected to possess enough generalization power to learn the correct system with appropriate physical meanings.

The performance and the quantitative results of the test MSE and the VPT are shown in Fig.~\ref{fig:pend_performance} and Table~\ref{tab:pend_table}, respectively. Note that it does not make sense to compare their training loss, since the definitions of the loss functions for SympNets and S-HNNs are quite different. Here, 10 independent experiments are simulated for each case to obtain the means and the standard deviations. We plot the results for the best models out of 10 in the first row of Fig.~\ref{fig:pend_performance}. LA-SympNets, with the smallest number of parameters, achieve the lowest prediction MSE and energy fluctuation.

Fig.~\ref{fig:pend_outer} shows that LA-SympNets generalize better than G-SympNets and S-HNNs on other trajectories. Given data on a single trajectory starting from $(0,1)$, LA-SympNets can learn the correct phase flow starting from $(0,1.5)$, $(0,2)$, $(0,2.5)$. It is worth mentioning that the prediction will deviate from the ground truth if the test trajectory goes farther away from the training data.

SympNets will be much easier to train when the training data is coarse-grained, or with large time steps. As mentioned before, since we do not assume the Hamiltonians are separable, one can only integrate $\Tilde{H}$ by implicit symplectic schemes, which means that only one-step methods can be used to train the S-HNNs. In general, high-order implicit symplectic schemes are not compatible with the HNN-type models and could take a much longer time to train. To make the comparison fair, we still use a one-step midpoint rule here as the integrator, but one can postulate, as the time step $h$ becomes larger, that the discretization error in S-HNNs would dominate and result in larger testing loss.

Here, we showcase a scenario when $h = 3$, which is roughly half of the period of the pendulum in our example. As shown in Table \ref{tab:pend_table}, S-HNNs fail to learn the correct dynamics of the system while SympNets continue to give the correct prediction for a long time period, according to the second row of Fig.~\ref{fig:pend_performance}.

\subsubsection{Learning irregularly sampled data} \label{subsubsec:learn_irr}
Here we make an extension of SympNets to learn the data with variable time steps. As aforementioned, a symplectic module can be written like
\begin{equation*}
    v(x) = \begin{bmatrix} I & f \\ 0 & I \end{bmatrix}(x)+b,
\end{equation*}
where $f$ could be $S$, $\tilde{\sigma}$ or $\hat{\sigma}$, depending on which type of module it belongs to, and with the bias $b$ being zero in the cases of activation module and gradient module.
We can insert a time step $h$ into the module as
\begin{equation*}
    v(x,h) = \begin{bmatrix} I & h\cdot f \\ 0 & I \end{bmatrix}(x)+h\cdot b.
\end{equation*}
Hence by extending each module in the constructed SympNets to the above form, we are able to feed the phase points $x$ with the time steps $h$ together as data into the extended SympNets $\psi(x,h)$ for training and testing. The results are shown in the third row of Fig.~\ref{fig:pend_performance} and Table~\ref{tab:pend_table}. LA, G and S-HNNs can all successfully learn the irregular data and give correct conserved energy. Specifically, we observe that SympNets perform better than S-HNNs, while LA-SympNets are slightly better than G-SympNets.

If the data used in HNN paper \citep{greydanus2019hamiltonian}, which includes time derivatives information are given, i.e., $\mathcal{T} = \{(x_i,\dot{x_i})\}_1^N$, then the extended SympNet $\psi(x,h)$ can also learn the fully-informed data $\mathcal{T}$ by optimizing the loss
\begin{equation*}
    MSE=\frac{1}{2d\cdot N}\sum_{i=1}^{N}\norm{\frac{\partial \psi}{\partial h}(x_i,0)-\dot{x_i}}^2.
\end{equation*}
This point could be further explored in the future. The capability of the extended SympNets on dealing with irregular data and fully-informed data indicates that SympNets can handle all the tasks that S-HNNs can handle, including learning the continuous time evolution of dynamics. It is certainly reasonable because the symplecticity of the phase flow encodes all the information of the dynamical system as a Hamiltonian system. Roughly speaking, $\psi(x,h)$ can be treated as a universal model representing the solution to an arbitrary Hamiltonian system.

\subsection{The Double Pendulum problem}
SympNets can readily handle the non-separable Hamiltonian systems, while S-HNNs should carefully choose the integrator if the Hamiltonian is non-separable. Here we consider a double pendulum system with the Hamiltonian \begin{equation*}
\begin{split}
     &H(p_1,p_2,q_1,q_2)\\
    =&\frac{m_2l_2^2p_1^2 + (m_1+m_2)l_1^2p_2^{2} - 2m_2l_1l_2p_1p_2\cos(q_1-q_2)}{2m_2l_1^2l_2^2(m_1+m_2\sin^2(q_1-q_2))}\\
    &-(m_1+m_2)gl_1\cos q_1 - m_2gl_2\cos q_2.
\end{split}
\end{equation*}
 The double pendulum system consists of a pendulum attached directly to another one. The $i$-th pendulum is made of a ball of mass $m_i$ connected to a massless rigid rod of length $l_i$, as is shown in Fig.~\ref{fig:illus_double_3b}. The motion of the system is driven by the local gravitational field $g$;  $q_i$ represents the angle of the $i$-th pendulum and $p_i$ represents its corresponding canonical momentum:
 \begin{equation*}
\begin{split}
    &p_1 = (m_1 + m_2) l_1^2 \dot{q_1} + m_2l_1l_2\dot{q_2}\cos(q_1 - q_2),\\
    &p_2 =  m_2 l_2^2 \dot{q_2} + m_2l_1l_2\dot{q_1}\cos(q_1 - q_2).
\end{split}
\end{equation*}
For simplicity we set $m_1 = m_2 = l_1 = l_2 = g = 1$.

\begin{figure*}[htbp]
    \centering
    \includegraphics[width=0.99\textwidth]{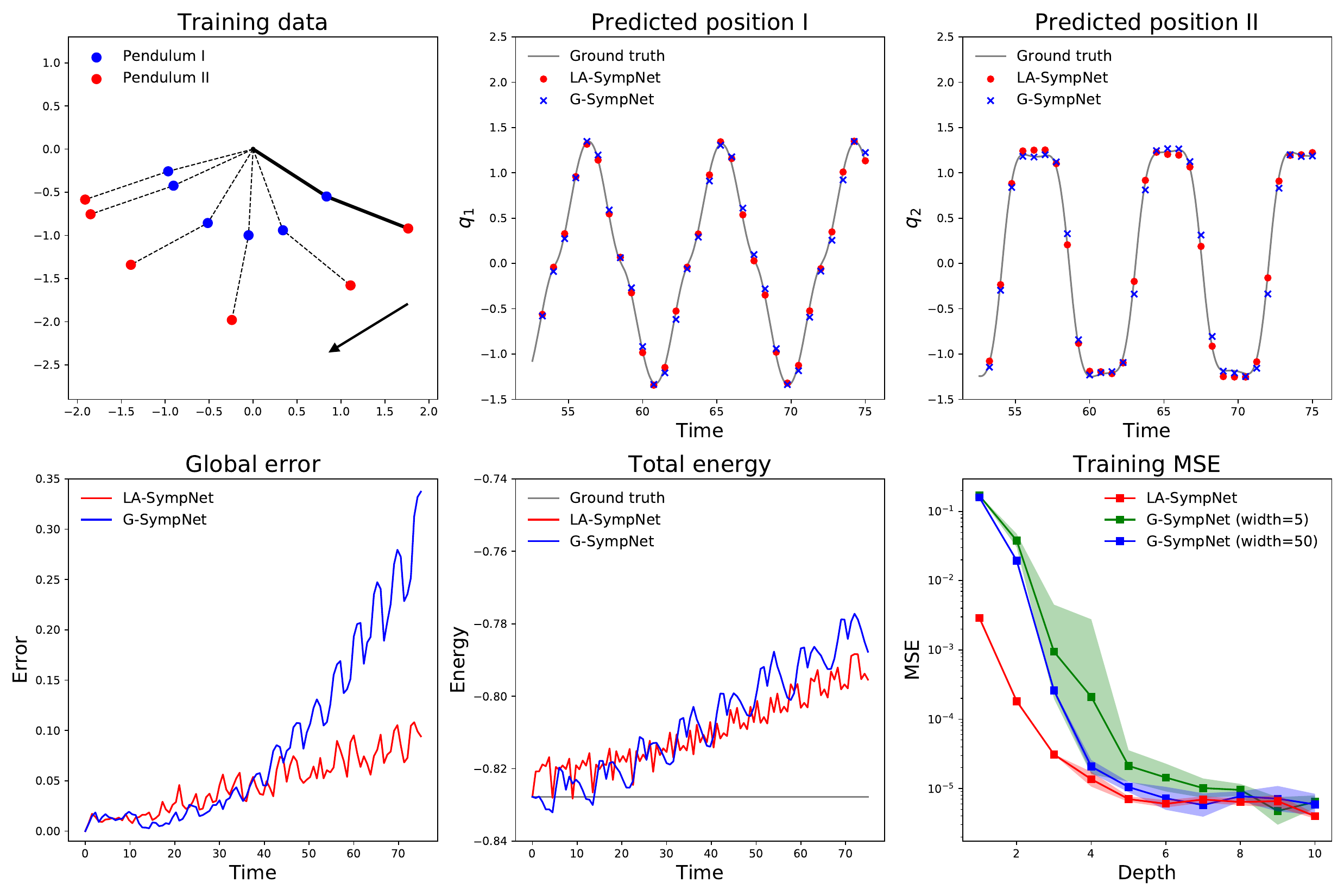}
    \caption{\textbf{Results for the double pendulum system.} (\textbf{Top-left}) Six consecutive points in the training dataset. The arrow represents the direction of motion. (\textbf{Top-middle and Top-right}) Predicted position $q$ for the two pendulums, respectively. The time window is chosen so that the difference between predictions made by LA/G-SympNets appear. (\textbf{Bottom-left}) The global error versus time. LA-SympNets generalize better than G-SympNets in the long term. (\textbf{Bottom-middle}) The total energies for the predicted trajectories. (\textbf{Bottom-right}) The training MSE versus the depth. The training MSE is obtained by taking the mean of 5 independent experiments, while the shaded region represents one standard deviation.}
    \label{fig:double_pendulum}
\end{figure*}
Similar to the pendulum example, the training dataset is made of $N = 200$ data points on a single trajectory starting from $x_0=(0,0,\frac{3\pi}{7},\frac{3\pi}{8})$ with time step $h = 0.75$.  The test dataset is given by the $k = 100$ data points following the last point in the training dataset, denoted by $X = (x_{N+1},\cdots,x_{N+k} )$.
The predictions made by SympNets are denoted by $\tilde{X} = (\Tilde{x}_{N+1},\cdots,\Tilde{x}_{N+k} )$. The mean squared error between $X$ and $\tilde{X}$ is taken as the test loss.

As shown in Fig.~\ref{fig:double_pendulum} and Table \ref{tab:double_three}, LA-SympNets outperform G-SympNets in the double pendulum prediction. The total energy of the trajectories predicted by SympNets matches the ground truth within a reasonable range. It is worth mentioning that S-HNNs completely fail in this task, because the time step $h = 0.75$ is so large that the discretization error in the numerical integrator dominates. This further demonstrates the advantage of SympNets when only sparsely sampled data is available. In contrast to the single pendulum case with $h = 3$, where one can remedy the
S-HNN by making the educated assumption that the Hamiltonian to be learned is separable, and discretize this system with high-order symplectic schemes, here the problem is more devastating since workable high-order symplectic methods for non-separable HNNs are in general more difficult to derive, and could result in intolerable computational expense.

According to the proofs of theorems \ref{thm:symp_LA} and \ref{thm:symp_G}, the approximation power of SympNets is determined by their depth and width (for G-SympNets). Indeed the bottom-right figure of Fig.~\ref{fig:double_pendulum} shows that the training MSE in this experiment decreases as the network grows deeper. Lower training errors are obtained for G-SympNets of width 50 compared to that of width 5, when the depth ranges from 1 to 8. However, the difference disappears when the depth becomes sufficiently large, which indicates the fact that depth plays a more important role than width for G-SympNets. Still, a wider network is preferred since the training process could become further stabilized. Among all the three models, the standard deviation of LA-SympNets is the lowest, demonstrating that LA-SympNets are more stable compared to G-SympNets.
\begin{table*}
\centering
\begin{tabular}{|c|c|c|c|}
\hline
Problem & LA-SympNet & G-SympNet & S-HNN \\ \hline
Double Pendulum  & $-3.4\pm 0.2$ & $-2.3\pm0.4$ & N/A \\ \hline
 Three-body & $-2.2\pm0.4$ & $-2.9\pm 0.2$ & $-1.8 \pm 0.4$ \\
 \hline
\end{tabular}
\caption{\textbf{The test loss for double pendulum and three-body.} The test loss is recorded in the form of mean $\pm$ standard deviation in log scale based on 10 independent experiments.}
\label{tab:double_three}
\end{table*}
\begin{figure*}[htbp]
    \centering
    \includegraphics[width=0.99\textwidth]{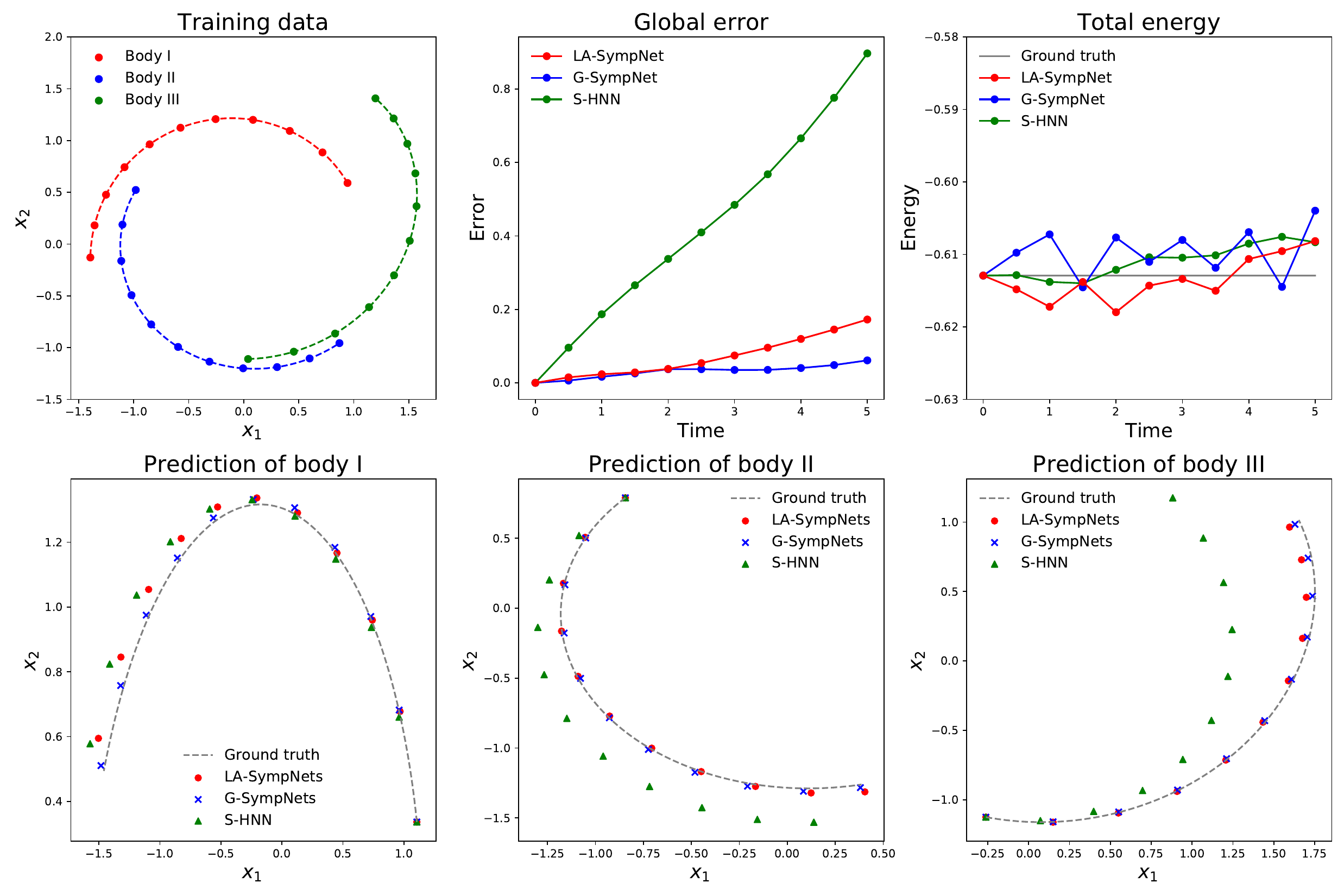}
    \caption{\textbf{Results for the three-body system.} (\textbf{Top-left}) One trajectory from the training dataset. The three position vectors $q_1$, $q_2$ and $q_3$ are plotted while the momentum vectors $p_1$, $p_2$ and $p_3$ are omitted for illustration purpose. (\textbf{Top-middle}) The global error versus time. The errors are calculated on one representative trajectory out of 1000. SympNets predict more accurately than S-HNNs on this and most of the other trajectories. (\textbf{Top-right}) The total energies for the predictions on the representative trajectory. (\textbf{Bottom}) Predicted position $q$ on the representative trajectory. SympNets can make predictions which stay on the true trajectory after a relatively longer time period.}
    \label{fig:three_body}
\end{figure*}
\subsection{The Three-Body problem}
To illustrate the fact that SympNets scale well to higher dimensions, we perform an experiment on the renowned three-body problem with a total number of 12 degrees of freedom. The Hamiltonian for this system is given by
\begin{equation*}
\begin{split}
    &H(\mathbf{p_1},\mathbf{p_2},\mathbf{p_3},\mathbf{q_1},\mathbf{q_2},\mathbf{q_3}) \\ =&\frac{\mathbf{p_1}^2}{2m_1} + \frac{\mathbf{p_2}^2}{2m_2} + \frac{\mathbf{p_3}^2}{2m_3}- \frac{Gm_1m_2}{|\mathbf{q_1}-\mathbf{q_2}|} - \frac{Gm_2m_3}{|\mathbf{q_2}-\mathbf{q_3}|} - \frac{Gm_1m_3}{|\mathbf{q_1}-\mathbf{q_3}|},
\end{split}
\end{equation*}
where $\mathbf{q_i} = (q_{i1}, q_{i2})$ represents the planar coordinates of the $i$-th body, while $\mathbf{p_i} = (p_{i1}, p_{i2})$ and $m_i$ are the corresponding momenta and mass, respectively; $G$ is the gravitational constant. For simplicity, we assume $G = m_1 = m_2 = m_3 = 1$.

Due to the chaotic nature of the system, it is almost impossible for a neural network model to make correct long-term predictions as in the pendulum case. So for both training and testing, we select $k = 10$ data points with time step $h = 0.5$ on each trajectory. In total, 5000 trajectories starting at random positions are simulated, among which 4000 are used as training data while the rest serve as test data, denoted by $\{X^{(i)}\}_{i=1}^{1000}$, where $X^{(i)} = (x^{(i)}_1, x^{(i)}_2, \cdots, x^{(i)}_{10})$. Similarly, the predictions are denoted by $\{\Tilde{X}^{(i)}\}_{i=1}^{1000}$. The average MSE between $X^{(i)}$ and $\Tilde{X}^{(i)}$ for all $1\leq i \leq 1000$ is taken as the test loss.

As can be seen from Table \ref{tab:double_three}, SympNets clearly outperform S-HNNs in terms of the test loss, while G-SympNets are slightly better that LA-SympNets in this task. Still, LA-SympNets are more memory-efficient in terms of their relatively smaller parameter size. The second row of Fig.~\ref{fig:three_body} shows that the two SympNet models are indeed comparable in their performances, while predictions made by S-HNNs completely fall off the trajectory, which is consistent with the results in \cite{greydanus2019hamiltonian}. All three methods are able to conserve the total energy of the Hamiltonian system.

\section{Summary}
The main contribution of this work is to provide a unified framework to infer dynamics from an arbitrary Hamiltonian system by utilizing the symplecticity of its phase flow. Just like any symplectic matrix that can be factorized into unit triangular matrices, in this paper we showed
that any symplectic map, which might be nonlinear, can be approximately factorized into unit triangular matrix-like maps in a simple form, i.e. SympNets. Furthermore, the SympNets are inherently reversible, and in fact form a group. This algebraic structure indicates the possibility of building normalizing flow models from the existing architecture. Besides its intriguing theoretical properties, SympNets also exhibit superior properties over competing baseline models, i.e., HNNs through the great performance in three numerical experiments including the pendulum, double pendulum and three-body problems. In particular, LA-SympNets generalize better than G-SympNets (pendulum, double pendulum), while G-SympNets are more expressive than LA-SympNets in more challenging scenarios (three-body problem). A new theoretical contribution is the universal approximation theorems (Theorem \ref{thm:symp_LA} and \ref{thm:symp_G}) that we proved for SympNets.

By constructing the SympNets, we wish our work could lead to more researches that focus on utilizing the underlying geometric structures such as the symplecticity in the data. In the future, we would like to derive generative models and control algorithms based on the SympNets. Another interesting direction will be to construct networks which could handle a larger class of systems including dissipative systems and systems with time-dependent Hamiltonians.

\section*{Acknowledgments}
The work of Pengzhan Jin, Aiqing Zhu and Yifa Tang was supported by the Major Project on New Generation of Artificial Intelligence from MOST of China (Grant No. 2018AAA0101002), and the National Natural Science Foundation of China (Grant No. 11771438). The work of Zhen Zhang and George Em Karniadakis was supported by the DOE PhILMs project (No. DE-SC0019453).


\appendix
\section{Proofs for universal approximation theorems} \label{app:proof_appro}
Some lemmas will be developed to prove these theorems.

Let $\Sigma_m[\sigma]$ denote the set of neural networks $f:\R^{m}\to \R$ with one hidden layer:
\begin{equation*}
\begin{split}
    \Sigma_m[\sigma]=\{&f(x)=a^T\sigma(Kx+b):\R^{m}\to \R|\\&a,b\in\R^n,K\in\R^{n\times m},n\in \mathbb{N}^*\},
\end{split}
\end{equation*}
where $\sigma$ is the activation function.
\begin{lemma}\label{lem:dev}
     $\Sigma_m[\sigma]$ is $r$-uniformly dense on compacta in $C^r(\R^{m})$ for $m\in \mathbb{N}^*$ if $\sigma$ is $r$-finite.
\end{lemma}
\begin{proof}
The proof can be found in \cite{HORNIK1990551}.
\end{proof}
Lemma \ref{lem:dev} indicates that neural networks with one hidden layer can approximate a function and its derivatives simultaneously, if the function satisfies certain regularity criteria.

\begin{lemma}\label{lem:gradv}
    Suppose $\sigma$ is $r$-finite, $V\in C^{r+1}(\R^{d})$, denote
    \begin{equation*}
        f\begin{pmatrix} p \\ q \end{pmatrix} = \begin{bmatrix}
        I & \nabla V  \\ 0 & I
        \end{bmatrix}\begin{pmatrix} p \\ q \end{pmatrix},\quad p,q\in \R^d,
    \end{equation*}
    where $\nabla V$ stands for the gradient of $V$, then for any compact $W\subset \R^{2d}$ and $\epsilon>0$, there exists $g\in \mathcal{M}_G$ such that $\norm{f-g}_{C^r(W;\R^{2d})}<\epsilon$.
\end{lemma}

\begin{proof}
Let $W_q = \{q\in\R^d|(p,q)\in W\}$. According to Lemma \ref{lem:dev}, there exists $\phi(x) = a^T(\int\sigma)(Kx+b)\in \Sigma_d[\int\sigma]$, such that $\norm{V-\phi}_{C^{r+1}(W_q;\R)}<\epsilon$, where $\int\sigma$ is $(r+1)$-finite. It can be seen that
\begin{equation*}
    \norm{\nabla V - \nabla \phi}_{C^r(W_q;\R^d)}\leq C\norm{V-\phi}_{C^{r+1}(W_q;\R)}<C\epsilon
\end{equation*}
for a positive constant $C$, which further implies
\begin{equation}\label{eqn:kak}
    \norm{\begin{bmatrix}I & \nabla V  \\ 0 & I \end{bmatrix}-\begin{bmatrix}I & \nabla \phi  \\ 0 & I \end{bmatrix}}_{C^r(W;\R^{2d})}=\norm{\nabla V - \nabla \phi}_{C^r(W_q;\R^d)}<C\epsilon.
\end{equation}
Note that $\nabla \phi(q) = K^Tdiag(a)\sigma(Kq+b)$ and let
\begin{equation*}
    g\begin{pmatrix} p \\ q \end{pmatrix}=\begin{bmatrix}I & \nabla \phi  \\ 0 & I \end{bmatrix}\begin{pmatrix} p \\ q \end{pmatrix}=\begin{pmatrix} p + K^Tdiag(a)\sigma(Kq+b) \\ q \end{pmatrix},
\end{equation*}
which is a gradient module in $\mathcal{M}_G$, then by (\ref{eqn:kak}), $\norm{f-g}_{C^r(W;\R^{2d})}<C\epsilon$.
\end{proof}
By symmetry, $f^*\begin{pmatrix} p \\ q \end{pmatrix} = \begin{bmatrix}
I & 0 \\ \nabla V & I
\end{bmatrix}\begin{pmatrix} p \\ q \end{pmatrix}$ can also be approximated by elements in $\mathcal{M}_G$ in the same way as in Lemma \ref{lem:gradv}. According to \cite{Turaev2002}, composition of Henon-like maps can approximate arbitrary symplectic maps. Thus, the problem reduces to approximate Henon-like maps by appropriate combination of $f$ and $f^*$.
\begin{definition}
The symplectic maps of the following form
\begin{equation*}
   \mathcal{H}[V]\begin{pmatrix} p \\ q \end{pmatrix}=\begin{bmatrix}
0 & I \\ -I & \nabla V
\end{bmatrix}\begin{pmatrix} p \\ q \end{pmatrix} = \begin{pmatrix} q \\ -p+\nabla V(q) \end{pmatrix}
\end{equation*}
for $V\in C^1(\R^d)$ are called Henon-like maps.
\end{definition}
\begin{lemma}\label{lem:hen}
Let $U\subset \R^{2d}$ be an open set, then for any $F\in \mathcal{SP}^r(U)$, compact $W\subset U$, $\epsilon>0$, there exists a sequence of $V_1,V_2,\cdots,V_n\in C^{r+1}(\R^d)$, such that
\begin{equation*}
    \norm{F - \mathcal{H}[V_n]\circ\cdots\circ \mathcal{H}[V_1]}_{C^r(W;\R^{2d})}<\epsilon
\end{equation*}
\begin{proof}
The proof can be found in \cite{Turaev2002}.
\end{proof}
\end{lemma}
\textbf{Proof of Theorem \ref{thm:symp_G}.} Let $W\subset U$ be a compact set, $\epsilon>0$. For $V\in C^{r+1}(\R^d)$, we have
\begin{equation*}
    \begin{bmatrix}0 & I \\ -I & \nabla V \end{bmatrix} = \begin{bmatrix}I & 0 \\ \nabla V  & I\end{bmatrix}\begin{bmatrix}I & I \\ 0 & I\end{bmatrix}\begin{bmatrix}I & 0 \\ -I & I\end{bmatrix}\begin{bmatrix}I & I \\ 0 & I\end{bmatrix},
\end{equation*}
which shows each $r$-th smooth Henon-like map can be represented as the composition of elements in
\begin{equation*}
    \mathcal{F}_G=\left\{\begin{bmatrix}I & \nabla V \\ 0 & I\end{bmatrix}\Bigg|V\in C^{r+1}(\R^d)\right\}\cup\left\{\begin{bmatrix}I & 0 \\ \nabla V  & I\end{bmatrix}\Bigg|V\in C^{r+1}(\R^d)\right\}.
\end{equation*}
According to Lemma \ref{lem:hen} and the above fact, any $F\in \mathcal{SP}^r(U)$ can be approximated by a sequence of $f_1,\cdots,f_n\in \mathcal{F}_G$ as
\begin{equation*}
    \norm{F - f_n\circ\cdots\circ f_1}_{C^r(W;\R^{2d})}<\epsilon.
\end{equation*}
Now we only need to prove the proposition: there exists $g_1,\cdots,g_n\in \mathcal{M}_G$ such that
\begin{equation*}
    \norm{f_n\circ\cdots\circ f_1 - g_n\circ\cdots\circ g_1}_{C^r(W;\R^{2d})}<\epsilon.
\end{equation*}

Denote
\begin{equation*}
    W_1=W,\quad W_i=T[f_{i-1}(W_{i-1})],\quad i=2,3,\cdots,n,
\end{equation*}
where the operator $T$ is defined as
\begin{equation*}
    T[A] = \{x\in \R^{2d}|\exists\ y\in A\ s.t.\ \norm{x-y}_\infty\leq 1\}
\end{equation*}
for compact $A\subset \R^{2d}$. It is easy to verify that $W_i$ is compact for $i=1, 2, 3, \cdots, n$. Let $\widetilde{W}=W_1\cup W_2\cup\cdots\cup W_n$. For $\epsilon_1,\cdots,\epsilon_n>0$, there exists $g_1,\cdots,g_{n}\in \mathcal{M}_G$ such that
\begin{equation*}
    \norm{f_i-g_i}_{C^{r}(\widetilde{W};\R^{2d})}<\min(\epsilon_i, 1),\quad i=1,2,\cdots,n,
\end{equation*}
according to Lemma \ref{lem:gradv}. With the definition of $g_i$, we know
\begin{equation*}
\begin{split}
    &\norm{g_i(x)-f_i(x)}_\infty \\
    \leq &\sup_{x\in \widetilde{W}}\norm{g_i(x)-f_i(x)}_\infty
    \leq \norm{f_i-g_i}_{C^{r}(\widetilde{W};\R^{2d})}
    <1,\quad \forall x\in W_i,
\end{split}
\end{equation*}
which derives that $g_i(W_i)\subset W_{i+1}$, hence we have
\begin{equation*}
\begin{split}
    &f_i\circ\cdots\circ f_1(W)\subset W_i\subset \widetilde{W},\\ &g_i\circ\cdots\circ g_1(W)\subset W_i\subset \widetilde{W},\quad i=1,\cdots,n.
\end{split}
\end{equation*}
Let $f_i=(f_i^{(1)},\cdots,f_i^{(2d)})^T$. Notice that
\begin{equation*}
    D^\alpha (f_n^{(k)}\circ f_{n-1}\circ\cdots\circ f_1)(x)=P_{\alpha,n,k}([D^\beta f_i^{(j)}(f_{i-1}\circ\cdots\circ f_1(x))]_{\beta,i,j})
\end{equation*}
which means $D^\alpha (f_n^{(k)}\circ f_{n-1}\circ\cdots\circ f_1)(x)$ can be represented as a polynomial depending on $\alpha,n,k$ with respect to the parameters $D^\beta f_i^{(j)}(f_{i-1}\circ\cdots\circ f_1(x))$ for $|\beta|\leq r, 1\leq i\leq n, 1\leq j \leq 2d$.

As $f_i\in C^r(\R^{2d};\R^{2d})\subset C^1(\R^{2d};\R^{2d})$, there holds Lipschitz condition on $\widetilde{W}$ for each $f_j\circ\cdots\circ f_{i+1}\circ f_i$ with a shared coefficient $L$:
\begin{equation*}
\begin{split}
    &\norm{f_j\circ\cdots\circ f_{i+1}\circ f_i(x)-f_j\circ\cdots\circ f_{i+1}\circ f_i(y)}_\infty \\ \leq &L\norm{x-y}_\infty,\quad \forall x,y\in \widetilde{W},\ \forall 1\leq i\leq j\leq n.
\end{split}
\end{equation*}
Then for any $x\in W$,
\begin{equation*}
\begin{split}
    &\norm{f_i\circ\cdots\circ f_1(x)-g_i\circ\cdots\circ g_1(x)}_\infty \\
    \leq &\sum_{k=1}^{i}\|f_i\circ\cdots f_{k+1}\circ f_k\circ g_{k-1}\circ\cdots\circ g_1(x) \\
    &- f_i\circ\cdots f_{k+1}\circ g_k\circ g_{k-1}\circ\cdots\circ g_1(x)\|_\infty \\
    \leq &L\cdot\left(\sum_{k=1}^{i-1}\norm{f_k\circ g_{k-1}\circ\cdots\circ g_1(x)-g_k\circ g_{k-1}\circ\cdots\circ g_1(x)}_\infty\right) \\
    &+\norm{f_i\circ g_{i-1}\circ\cdots\circ g_1(x)-g_i\circ g_{i-1}\circ\cdots\circ g_1(x)}_\infty \\
    \leq &\max(L,1)\cdot\left(\sum_{k=1}^{i}\norm{f_k-g_k}_{C^r(\widetilde{W};\R^{2d})}\right) \\
    \leq &\max(L,1)(\sum_{k=1}^{i}\epsilon_k)\leq \max(L,1)(\sum_{k=1}^{n}\epsilon_k). \\
\end{split}
\end{equation*}
Since $D^\beta f_i^{(j)}$ are uniformly continuous on $\widetilde{W}$,
\begin{equation*}
\begin{split}
\lim_{\epsilon_1,\cdots,\epsilon_n\to 0}\sup_{x\in W}&|D^\beta f_i^{(j)}(f_{i-1}\circ\cdots\circ f_1(x))\\
&-D^\beta f_i^{(j)}(g_{i-1}\circ\cdots\circ g_1(x))|=0,
\end{split}
\end{equation*}
consequently
\begin{equation*}
\begin{split}
\lim_{\epsilon_1,\cdots,\epsilon_n\to 0}\sup_{x\in W}&|D^\beta f_i^{(j)}(f_{i-1}\circ\cdots\circ f_1(x))\\
&-D^\beta g_i^{(j)}(g_{i-1}\circ\cdots\circ g_1(x))|=0,
\end{split}
\end{equation*}
due to
\begin{equation*}
\begin{split}
    &|D^\beta f_i^{(j)}(f_{i-1}\circ\cdots\circ f_1(x))-D^\beta g_i^{(j)}(g_{i-1}\circ\cdots\circ g_1(x))| \\
    \leq &|D^\beta f_i^{(j)}(f_{i-1}\circ\cdots\circ f_1(x))-D^\beta f_i^{(j)}(g_{i-1}\circ\cdots\circ g_1(x))|\\
    &+\norm{f_i-g_i}_{C^r(\widetilde{W};\R^{2d})} \\
    < &|D^\beta f_i^{(j)}(f_{i-1}\circ\cdots\circ f_1(x))-D^\beta f_i^{(j)}(g_{i-1}\circ\cdots\circ g_1(x))|\\
    &+ \epsilon_i.
\end{split}
\end{equation*}
Therefore,
\begin{equation*}
\begin{split}
    &\lim_{\epsilon_1,\cdots,\epsilon_n\to 0}\norm{f_n\circ\cdots\circ f_1 - g_n\circ\cdots\circ g_1}_{C^r(W;\R^{2d})} \\
    =&\lim_{\epsilon_1,\cdots,\epsilon_n\to 0}\sum_{|\alpha|\leq r}\max_{1\leq k\leq n}\sup_{x\in W} |D^\alpha (f_n^{(k)}\circ f_{n-1}\circ\cdots\circ f_1)(x)\\
    &-D^\alpha (g_n^{(k)}\circ g_{n-1}\circ\cdots\circ g_1)(x)| \\
    =&\sum_{|\alpha|\leq r}\max_{1\leq k\leq n}\lim_{\epsilon_1,\cdots,\epsilon_n\to 0}\sup_{x\in W} |P_{\alpha,n,k}([D^\beta f_i^{(j)}(f_{i-1}\circ\cdots\circ f_1(x))]_{\beta,i,j}) \\
    &-P_{\alpha,n,k}([D^\beta g_i^{(j)}(g_{i-1}\circ\cdots\circ g_1(x))]_{\beta,i,j})| \\
    =& 0,
\end{split}
\end{equation*}
where the last equal holds since $D^\beta f_i^{(j)}(f_{i-1}\circ\cdots\circ f_1(W))$ and $D^\beta g_i^{(j)}(g_{i-1}\circ\cdots\circ g_1(W))$ are uniformly bounded in a larger compact set for all $\beta,i,j$ as $\{\epsilon_i\}\to 0$, as well as $P_{\alpha,n,k}$ is uniformly continuous on that bounded compact set. Hence, the proposition has been completed. \hfill\qedsymbol

\textbf{Proof of Theorem \ref{thm:symp_LA}.} Recall that
\begin{equation*}
\begin{split}
    \mathcal{M}_G=&\Bigg\{\begin{bmatrix} I & \hat{\sigma}_{K,a,b} \\ 0 & I \end{bmatrix}\Bigg|
    K\in \R^{n\times d},a,b\in \R^{nd},n\in\mathbb{N}^*\Bigg\} \\
    &\cup \Bigg\{\begin{bmatrix} I & 0 \\ \hat{\sigma}_{K,a,b} & I \end{bmatrix}\Bigg|K\in \R^{n\times d},a,b\in \R^{nd},n\in\mathbb{N}^*\Bigg\},
\end{split}
\end{equation*}
Here we rewrite $\mathcal{M}_G$ in a slightly different form as
\begin{equation*}
\begin{split}
    \mathcal{M}_G=&\Bigg\{\begin{bmatrix} I & \hat{\sigma}_{K,a,b} \\ 0 & I \end{bmatrix}\Bigg|
    K\in \R^{nd\times d},a,b\in \R^{nd},n\in\mathbb{N}^*\Bigg\} \\
    &\cup \Bigg\{\begin{bmatrix} I & 0 \\ \hat{\sigma}_{K,a,b} & I \end{bmatrix}\Bigg|K\in \R^{nd\times d},a,b\in \R^{nd},n\in\mathbb{N}^*\Bigg\},
\end{split}
\end{equation*}
by extending $K,a,b$ with some zero rows to meet the requirement of width being multiple of $d$. Furthermore, denote
\begin{equation*}
\begin{split}
    &\mathcal{K}_n \\
    =&\{K\in \R^{nd\times d}|K=(K_1^T,\cdots,K_n^T)^T,K_i\in \R^{d\times d}, \det(K_i)\neq 0\},
\end{split}
\end{equation*}
\begin{equation*}
\begin{split}
    \widetilde{\mathcal{M}}_G=&\Bigg\{\begin{bmatrix} I & \hat{\sigma}_{K,a,b} \\ 0 & I \end{bmatrix}\Bigg|
    K\in \mathcal{K}_n,a,b\in \R^{nd},n\in\mathbb{N}^*\Bigg\} \\
    &\cup \Bigg\{\begin{bmatrix} I & 0 \\ \hat{\sigma}_{K,a,b} & I \end{bmatrix}\Bigg|K\in \mathcal{K}_n,a,b\in \R^{nd},n\in\mathbb{N}^*\Bigg\},
    \end{split}
\end{equation*}
and
\begin{equation*}
    \widetilde{\Psi}_G=\{\psi=u_k\circ\cdots\circ u_1|u_i\in \widetilde{\mathcal{M}}_G, k\in \mathbb{N}^*\}\subset \Psi_G.
\end{equation*}
Given any compact set $W\subset U$ and $\psi_{\{K_i,a_i,b_i\}}=u_{K_k,a_k,b_k}\circ\cdots\circ u_{K_1,a_1,b_1}\in \Psi_G$, we can easily verify that
\begin{equation*}
    \lim_{\{\widetilde{K}_i\}\to \{K_i\}}\norm{\psi_{\{K_i,a_i,b_i\}}-\psi_{\{\widetilde{K}_i,a_i,b_i\}}}_{C^r(W;\R^{2d})}=0,\quad\psi_{\{\widetilde{K}_i,a_i,b_i\}}\in \widetilde{\Psi}_G,
\end{equation*}
since $\mathcal{K}_n$ is dense in $\R^{nd\times d}$ and \begin{equation*}
    f(\{\widetilde{K}_i\})=\norm{\psi_{\{K_i,a_i,b_i\}}-\psi_{\{\widetilde{K}_i,a_i,b_i\}}}_{C^r(W;\R^{2d})}
\end{equation*} is continuous with respect to $\{\widetilde{K}_i\}$. Therefore $\widetilde{\Psi}_G$ is $r$-uniformly dense on compacta in $\Psi_G$, furthermore, is $r$-uniformly dense on compacta in $\mathcal{SP}^r(U)$ by Theorem \ref{thm:symp_G}.

On the other hand, given
\begin{equation*}
\begin{split}
    &\phi\begin{pmatrix} p \\ q \end{pmatrix}=\begin{pmatrix} p \\ K^Tdiag(a)\sigma(Kp+b)+q \end{pmatrix}\in \widetilde{\mathcal{M}}_G, \\
    &K=(K_1^T,\cdots,K_n^T)^T\in\mathcal{K}_n, \\
    &a=(a_1^T,\cdots,a_n^T)^T,a_i\in \R^{d}, \\
    &b=(b_1^T,\cdots,b_n^T)^T,b_i\in \R^{d},
\end{split}
\end{equation*}
define
\begin{equation*}
\begin{split}
    v_i\begin{pmatrix}p \\ q\end{pmatrix}=&\begin{pmatrix}K_i^{-1} & 0  \\ 0 & K_i^{T}\end{pmatrix}\begin{bmatrix}I & 0 \\ diag(a_i)\sigma & I\end{bmatrix} \\
    &\left(\begin{pmatrix}K_i & 0  \\ 0 & K_i^{-T}\end{pmatrix}\begin{pmatrix}p \\ q\end{pmatrix}+\begin{pmatrix}b_i \\ 0\end{pmatrix}\right)-\begin{pmatrix}K_i^{-1}b_i \\ 0\end{pmatrix}
\end{split}
\end{equation*}
for $i=1,\cdots,n$. Theorem \ref{thm:repre_thm} points out that $v_i\in \Psi_{LA}$, and one may readily check that $\phi=v_n\circ\cdots\circ v_1$. Subsequently, we know $\widetilde{\Psi}_G\subset\Psi_{LA}$, thus $\Psi_{LA}$ is $r$-uniformly dense on compacta in $\mathcal{SP}^r(U)$.\hfill\qedsymbol

\bibliographystyle{elsarticle-harv}
\bibliography{main}





\end{document}